\documentclass[review,3p,times]{elsarticle}

\usepackage{amssymb}
\usepackage{latexsym}
\usepackage{graphicx}
\usepackage{amsfonts,amssymb,amsmath}
\usepackage{hyperref}
\usepackage{adjustbox}
\usepackage{array}
\usepackage{multirow,booktabs}
\usepackage[utf8]{inputenc}
\usepackage[english]{babel}
\usepackage[table]{xcolor}
\usepackage{caption} 
\usepackage{ulem}
\usepackage{subcaption}
\usepackage{longtable}
\usepackage{tabularx}
\usepackage{ltablex}

\journal{Knowledge-Based System}

\begin{document}

\begin{frontmatter}



\title{Table Detection for Visually Rich Document Images}


\author[inst1]{Bin Xiao}
\ead{bxiao103@uottawa.ca}
\author[inst1]{Murat Simsek}
\ead{murat.simsek@uottawa.ca}
\author[inst1]{Burak Kantarci}
\ead{burak.kantarci@uottawa.ca}

\affiliation[inst1]{organization={School of Electrical Engineering and Computer Science},
            addressline={University of Ottawa}, 
            city={Ottawa},
            postcode={K1N 6N5}, 
            state={ON},
            country={Canada}}

\author[inst2]{Ala Abu Alkheir}
\ead{ala\_abualkheir@lytica.com}

\affiliation[inst2]{organization={Lytica Inc.},
             addressline={555 Legget Drive}, 
            city={Ottawa},
            postcode={K2K 2X3}, 
            state={ON},
            country={Canada}}

\tnotetext[myNSFnote]
{
This work was supported in part by Mathematics of Information Technology and Complex Systems (MITACS) Accelerate Program, Smart Computing for Inovation (SOSCIP) Program and Lytica Inc.
} 

\begin{abstract}

Table Detection (TD) is a fundamental task to enable visually rich document understanding, which requires the model to extract information without information loss. However, popular Intersection over Union (IoU) based evaluation metrics and IoU-based loss functions for the detection models cannot directly represent the degree of information loss for the prediction results. Therefore, we propose to decouple IoU into a ground truth coverage term and a prediction coverage term, in which the former can be used to measure the information loss of the prediction results. Besides, considering the sparse distribution of tables in document images, we use SparseR-CNN as the base model and further improve the model by using Gaussian Noise Augmented Image Size region proposals and many-to-one label assignments. Results under comprehensive experiments show that the proposed method can consistently outperform state-of-the-art methods with different IoU-based metrics under various datasets and demonstrate that the proposed decoupled IoU loss can enable the model to alleviate information loss.

\end{abstract}




\end{frontmatter}


\section{Introduction}
\label{sec:introduction}

Table Detection (TD) is often a pre-processor step for information extraction and document understanding tasks~\cite{xiao2022handling, xiao2022efficient}. One typical formulation for the TD problem is transforming the electronic documents into images and then using object detection models to generate the bounding boxes of defined table objects. Current state-of-the-art methods~\cite{fernandes2022tabledet, prasad2020cascadetabnet} for the TD problem usually employ two-stage object detectors, which require dense candidates and apply data augmentation and multiple-stage transfer learning techniques. However, tables in visually rich documents are usually well formatted and large so that human readers can easily interpret them. Besides, the number of tables in a single document image is typically small, which means their distribution in a single document is sparse. Based on these observations, we use SparseR-CNN~\cite{sun2021sparse} as the base model in this study, which is a competitive detector using sparse learnable regional proposals. It is worth mentioning that many state-of-the-art studies for the TD problem often employ two-stage detectors using dense candidates and multiple-stage transfer learning techniques, which are usually more complex than the proposed method. We also propose to use image size regional proposals to cover all the information of target tables in the proposal boxes and use the noise augmentation method to enrich the diversity of proposal boxes.

\begin{figure}[htp]
\begin{center}
 \includegraphics[width=1.0\columnwidth]{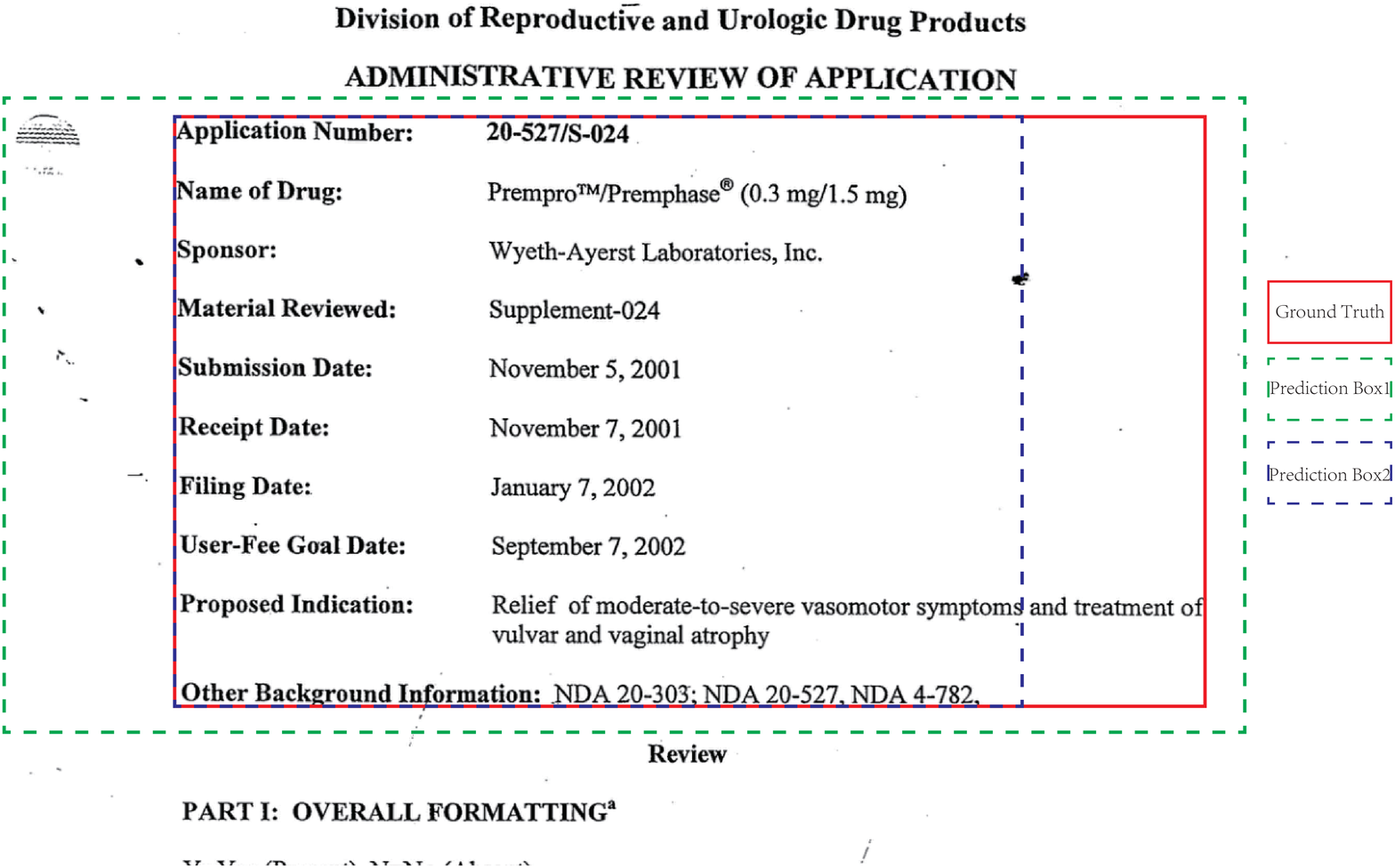}
  \caption{Preferable prediction for TD tasks. The IoU scores of green and blue predictions are 0.77 and 0.82, respectively. The green prediction is preferable for TD tasks, even though its IoU score is smaller.}
  \label{fig:box_perference}
\end{center}
\end{figure}

Since information extraction tasks often follow TD tasks, it is vital to avoid information loss. Therefore, a larger prediction box is preferable to a smaller box that can lose information even when the latter box has a larger Intersection over Union (IoU) score with the ground truth. Figure~\ref{fig:box_perference} uses a table as an example to further illustrate this observation. The green box in Figure~\ref{fig:box_perference} has an IoU score of 0.77 with the red ground truth box, while the blue box has an IoU score of 0.82. Even though the blue prediction has a larger IoU score, the green prediction is preferable for TD tasks. Motivated by these observations, we argue that the IoU score cannot directly reflect the information loss of a prediction box. Therefore, we propose to decouple the IoU score into two terms: a ground truth coverage term and a prediction coverage term, in which the former term can be used to measure the information loss for the prediction boxes. It is worth mentioning that the proposed decoupled IoU score termed the Information Coverage Score (ICS), can replace the IoU score in the IoU-based loss functions and evaluation metrics.

\begin{figure}[htp]
\begin{center}
  \includegraphics[width=1.0\columnwidth]{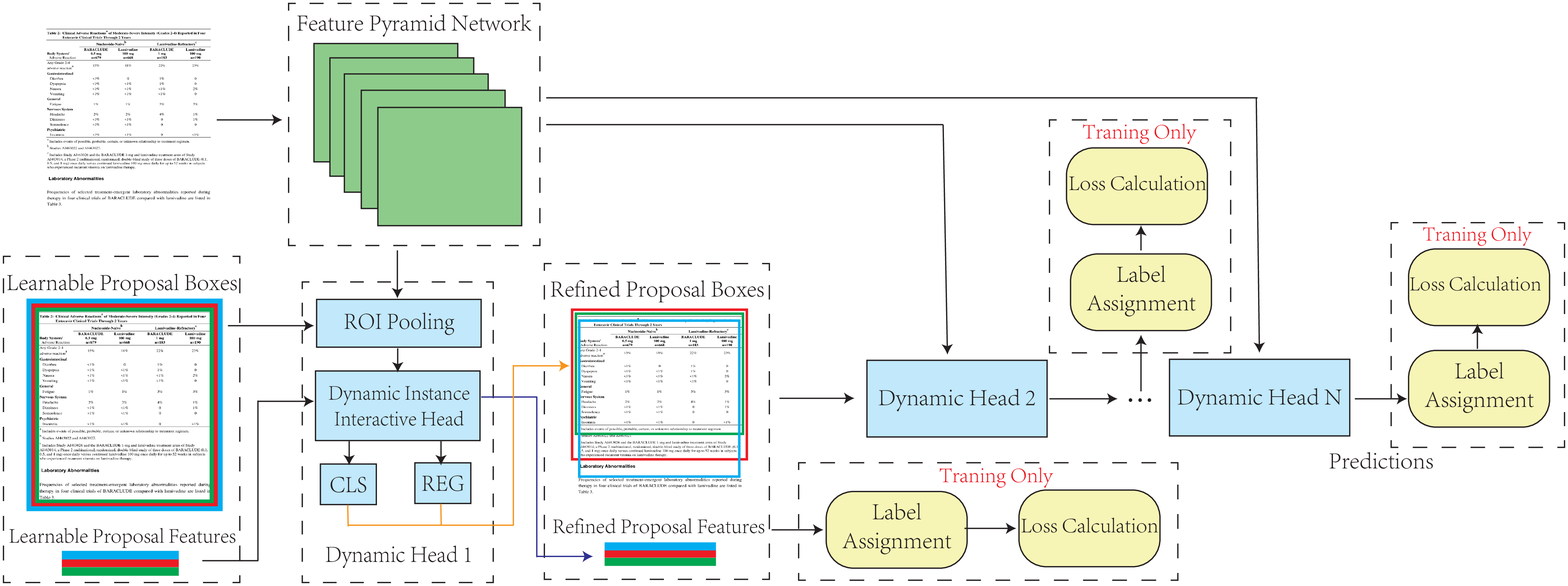}
  \caption{The overall architecture of the proposed method. Notably, all Dynamic Heads share an identity structure. We only show the details of Dynamic Head1 in this figure for simplicity.}
  \label{fig:architecture}
\end{center}
\end{figure}

On the other hand, label assignment in object detection models is to define the classification and regression targets for anchors~\cite{ge2021ota}. Many studies~\cite{hong2022dynamic, ge2021ota} have shown that label assignment plays a vital role in the success of a detector, and the one-to-one scheme used in SparseR-CNN~\cite{sun2021sparse} is not optimal. Besides, SparseR-CNN employs a cascade architecture that uses the outputs of $i$th Dynamic Head as the inputs of the $i+1$th Dynamic Head to refine the predictions, as shown in Figure~\ref{fig:architecture}, which means that the proposal quality of each Dynamic Head varies. Therefore, inspired by the studies~\cite{hong2022dynamic, chen2022diffusiondet, ge2021ota, du20211st, ge2021yolox}, we leverage a SimOTA~\cite{ge2021yolox} based many-to-one label assignment approach, which further improves the SimOTA by adapting a dynamic scheduling scheme to adjust the number of positive assignments dynamically and integrating the proposed ICS loss to the cost function.

To sum up, the contributions of this study are three-fold:
\begin{enumerate}

\item We introduce a decoupled IoU score, termed Information Coverage Score (ICS), which can reflect the information loss of the prediction boxes when it is used as evaluation metrics and encourage the model to alleviate the information loss when it is used as loss functions.

\item We improve the SimOTA method by adapting a dynamic scheduling scheme and integrating the ICS loss, propose a Gaussian Noise Augmented Image Size region proposal method, and apply them to the SpareR-CNN model to further improve the performance of the proposed method.

\item To compare with state-of-the-art models fairly, we first conduct extensive experiments using IoU-based evaluation metrics and loss functions on various manually annotated datasets to demonstrate the efficiency and effectiveness of our proposed detection model. Then, we conduct further experiments to demonstrate the benefits of the proposed ICS score when it is applied to the TD problem. The experimental results show that the proposed method can consistently outperform the state-of-the-art benchmark models under different evaluation metrics. 

\end{enumerate}

The rest of this paper is organized as follows: Section~\ref{sec:related_work} discusses related studies, including studies in Object Detection models and Table Detection models. Section~\ref{sec:proposed_method} introduces the formal problem definition and our proposed SparseTableDet model. Section~\ref{sec:experiments} presents the experiments and discusses the design aspects of the proposed method. At last, we draw our conclusion and possible directions in section~\ref{sec:conclusion}.


\section{Related Work}
\label{sec:related_work}
As discussed in section~\ref{sec:introduction}, we formulate the TD problem as an Object Detection problem and propose an ICS that can replace the IoU in loss functions and evaluation metrics. Therefore, this section first discusses popular object detection models and loss functions. Since the proposed method employs a Many-to-One label assignment, we include popular label assignment studies. At last, we focus on the studies specifically designed or optimized for the TD problem.

\subsection{Object Detection Methods}
\label{sec:object_detection_models}

Object detection problem has been widely discussed in recent years. Object detection models are often categorized into one-stage and two-stage models based on their number of regression steps. Popular one-stage models include YOLO~\cite{redmon2016you} and its variants ~\cite{shafiee2017fast, redmon2018yolov3, bochkovskiy2020yolov4, ge2021yolox}, SSD~\cite{liu2016ssd}, FCOS~\cite{tian2019fcos}, and many others. These one-stage models do not contain the step of generating region proposals and usually have faster training and inference speed compared with two-stage models. In contrast, two-stage models usually first use a region proposal network to generate a series of regional proposals, then further regress and classify the regional proposals by well-designed models. Typical two-stage models include FasterR-CNN~\cite{ren2015faster}, MaskR-CNN~\cite{he2017mask}, CascadeR-CNN~\cite{cai2018cascade} and many others. Along with these typical one-stage and two-stage models, transformer-based approaches recently attracted a lot of attention. DETR~\cite{carion2020end} is the first study introducing transformer and self-attention mechanism ~\cite{vaswani2017attention} to the object detection. Following the design of DETR, many variants of DETR have been proposed to improve the performance further and accelerate the convergence, such as Deformable-DETR~\cite{zhu2020deformable} and DAB-DETR~\cite{liu2022dab}. SparseR-CNN~\cite{sun2021sparse} adopts the Set prediction loss and Hungarian matching from DETR and proposes to use learnable region proposals and learnable instance features to simplify the region proposal generation. One key characteristic of the learnable region proposals in SparseR-CNN is that they can be initialized with different methods, including random initialization and image size initialization. Our proposed method in this study is based on SparseR-CNN and uses the image size initialization considering the requirements of TD applications as discussed in section~\ref{sec:introduction}.

Loss functions used in object detectors' regression and classification tasks have also been discussed widely. For the regression task, it is a natural choice to use $l_n$-norms and their variants, such as smooth-l1~\cite{girshick2015fast}, as the loss function. However, these functions are not aligned with the widely accepted evaluation metric IoU score, meaning that for some cases minimizing these loss functions cannot lead to better IoU scores~\cite{rezatofighi2019generalized, zheng2022scaloss}. IoU-based loss functions can alleviate this issue and become the most popular choice. IoU loss~\cite{yu2016unitbox} has the gradient vanishing issue when the prediction and ground truth boxes have no overlaps. GIoU loss ~\cite{rezatofighi2019generalized} extends the IoU loss by adding an extra penalty term to alleviate the gradient vanishing problem when two boxes have no overlap. More specifically, assuming that $A$, $B$ denote two arbitrary convex shapes and $C$ is the smallest enclosing convex, then the term used in GIoU loss is defined as $\frac{|C - A \cup B|}{|C|}$, where $-$ means the complementary operation. A limitation of GIoU loss is that it can be degraded to IoU loss for enclosing bounding boxes. To address this limitation of GIoU loss, DIoU loss~\cite{zheng2020distance} proposes to use the distance between two boxes' centers as the additional term, which can lead to faster convergence and alleviate the gradient vanishing problem. CIoU loss~\cite{zheng2021enhancing} also considers the geometric factors of bounding boxes and proposes an aspect ratio term, a distance term, and the IoU term. Besides these popular IoU-based loss functions, there are other loss functions without using IoU score, such as SCALoss~\cite{zheng2022scaloss} and KLLoss~\cite{he2019bounding}. SCALoss defines two terms considering side overlap and corner distance. KLLoss requires the output to be a distribution instead of location coordinates. 

Besides these object detection models, label assignment methods have also been widely discussed in recent studies. Typically, label assignment methods can be categorized into Fixed Label Assignment and Dynamic Label Assignment~\cite{ge2021ota}. Fixed Label Assignments are methods defining fixed criterion to determine the positive and negative samples of each ground truth. For example, Region Proposal Network (RPN) in FasterR-CNN~\cite{ren2015faster} defines two IoU scores as the thresholds to the positive and negative proposals. YOLO~\cite{redmon2016you} uses the closest anchor points to the center of ground truths as positive anchor points. In contrast, Dynamic Label Assignment methods often formulate the problem as an optimization problem and solve the problem more dynamically. For example, OTA~\cite{ge2021ota} formulates the label assignment problem as an Optimal Transport (OT) problem, which can be optimized by the Sinkhorn-Knopp algorithm. SimOTA~\cite{du20211st, ge2021yolox}  uses the top-k candidates whose centers are in the ground truth bounding boxes to avoid the time-consuming optimization process.

\subsection{Table Detection}
\label{sec:table_detection}
Many studies have discussed the TD problem recently. One of the most popular formulations for the TD problem is defining tables in visually rich documents as objects and then applying popular object detectors. Following this problem formulation, the object detection approaches discussed in section~\ref{sec:object_detection_models} can be easily adapted to the TD problem and widely used as benchmark models in many studies~\cite{abdallah2022tncr}. Considering the special characters and requirements for the TD problem, many studies further optimized the popular object detection methods to improve the model performance. Due to the limited number of training samples for the TD problem, transfer learning methods are widely used. CascadeTabNet~\cite{prasad2020cascadetabnet} extends the Cascade Mask R-CNN~\cite{cai2018cascade} model and uses HRNet~\cite{wang2020deep} as the backbone network. Besides, CascadeTabeNet applies a two-stage transfer learning approach and various augmentation methods. Similarly, TableDet~\cite{fernandes2022tabledet} is based on Cascade R-CNN~\cite{cai2018cascade}, proposes a Table Aware Cutout augmentation method, and also leverages a two-step transfer learning approach to improve the model performance further. Besides two-stage detectors, one-stage methods, such as YOLO~\cite{redmon2016you} and its variants, also have been adapted to the TD problem. YOLOv3-TD~\cite{huang2019yolo} employs the YOLOv3~\cite{redmon2018yolov3} as the base model and proposes some adaptive adjustments, including a new anchor optimization method and a new post-processing process. Besides these one-stage and two-stage methods, transformer-based approaches such as DETR~\cite{carion2020end} and Deformable-Detr~\cite{zhu2020deformable}, also have been applied to the TD problem~\cite{smock2022pubtables, abdallah2022tncr}. There are many other studies discussing the TD problem, including DeepDeSRT~\cite{schreiber2017deepdesrt}, TableDet~\cite{fernandes2022tabledet}, and many others~\cite{siddiqui2018decnt, kara2020holistic}. All in all, these studies usually adopt popular object detection models to the TD problem and use some specifically designed methods to improve the model performance based on the characteristics of the TD problem.

\section{Proposed Method}
\label{sec:proposed_method}
\subsection{Architecture of the Proposed Method}

Following the architecture of SparseR-CNN~\cite{sun2021sparse}, our proposed method also consists of an Initialization Module, a Feature Pyramid Network, and a series of Dynamic Heads. The Initialization Module is used to initialize the learnable proposal boxes and the learnable proposal features. In this study, we use the Noise Augmented Image Size region proposals, which will be discussed in section~\ref{sec:noise_augment_to_proposals}. Feature Pyramid Network (FPN)~\cite{lin2017feature} is the backbone network to generate image features for every Dynamic Head. The Dynamic Heads are used to do the regression and classification tasks. Dynamic Head $t+1$ takes the image features generated by FPN and the outputs of Dynamic Head $t$ , including the Refined Proposal Features and Refined Proposal Boxes, as the input to further refine the predictions of Dynamic Head $t$ when $t > 1$. Since the predictions of each Dynamic Head are used to calculate the loss, a label assignment process is operated on these predictions to further calculate the losses. Since our refinements to the SparseR-CNN model are mainly on the proposal initialization, label assignment, and the loss functions, which will be discussed in section~\ref{sec:noise_augment_to_proposals}, \ref{sec:many_to_one_label_assignment} and \ref{sec:information_coverage_score}, we keep the default implementations of SparseR-CNN for other parts which are detailed described in the study~\cite{sun2021sparse}.

\subsection{Noise Augmentation to Region Proposals}
\label{sec:noise_augment_to_proposals}
As discussed in section~\ref{sec:introduction}, TD applications typically require predictions to avoid information loss, and the tables in the documents are usually large and have no overlaps. Considering these characteristics of TD applications, using Image Size to initialize the region proposals becomes a good choice compared with other initialization methods, such as Random Initialization~\cite{sun2021sparse} and Grid Initialization~\cite{sun2021sparse}, because it can avoid information loss at the first step of the detector. However, simply using a number of the same proposals may not be optimal. Therefore, we propose a simple but effective augmentation method to the region proposals by adding Gaussian Noise to enrich the proposals' diversity. More specifically, assuming that a proposal box is represented by its box center, width, and height, namely $b = \{c_x, c_y, w, h\}$, then the augmented proposal box can be defined as Equation~\ref{eq:proposal_box_augmentation}, where $\mathcal{N}$ means Gaussian Distribution. In our implementation, boxes are normalized, meaning that an image size box can be represented as $b = \{0.5, 0.5, 1, 1\}$, and $\mu$, $\sigma^2$ are set as 0 and 0.01, respectively. 

\begin{equation}
\label{eq:proposal_box_augmentation}
\begin{aligned}
 b_{aug} = f (\{c_x, c_y, w, h\}) =  \{c_x + \epsilon_x, c_y + \epsilon_y, w - 2 \cdot |\epsilon_x|, h - 2 \cdot |\mathcal{\epsilon}_y| \}, \epsilon_x \in \mathcal{N}(\mu,\,\sigma^{2}), \epsilon_y \in \mathcal{N}(\mu,\,\sigma^{2})
\end{aligned}
\end{equation}

\begin{figure}[htp]
\begin{center}
  \includegraphics[width=0.4\columnwidth]{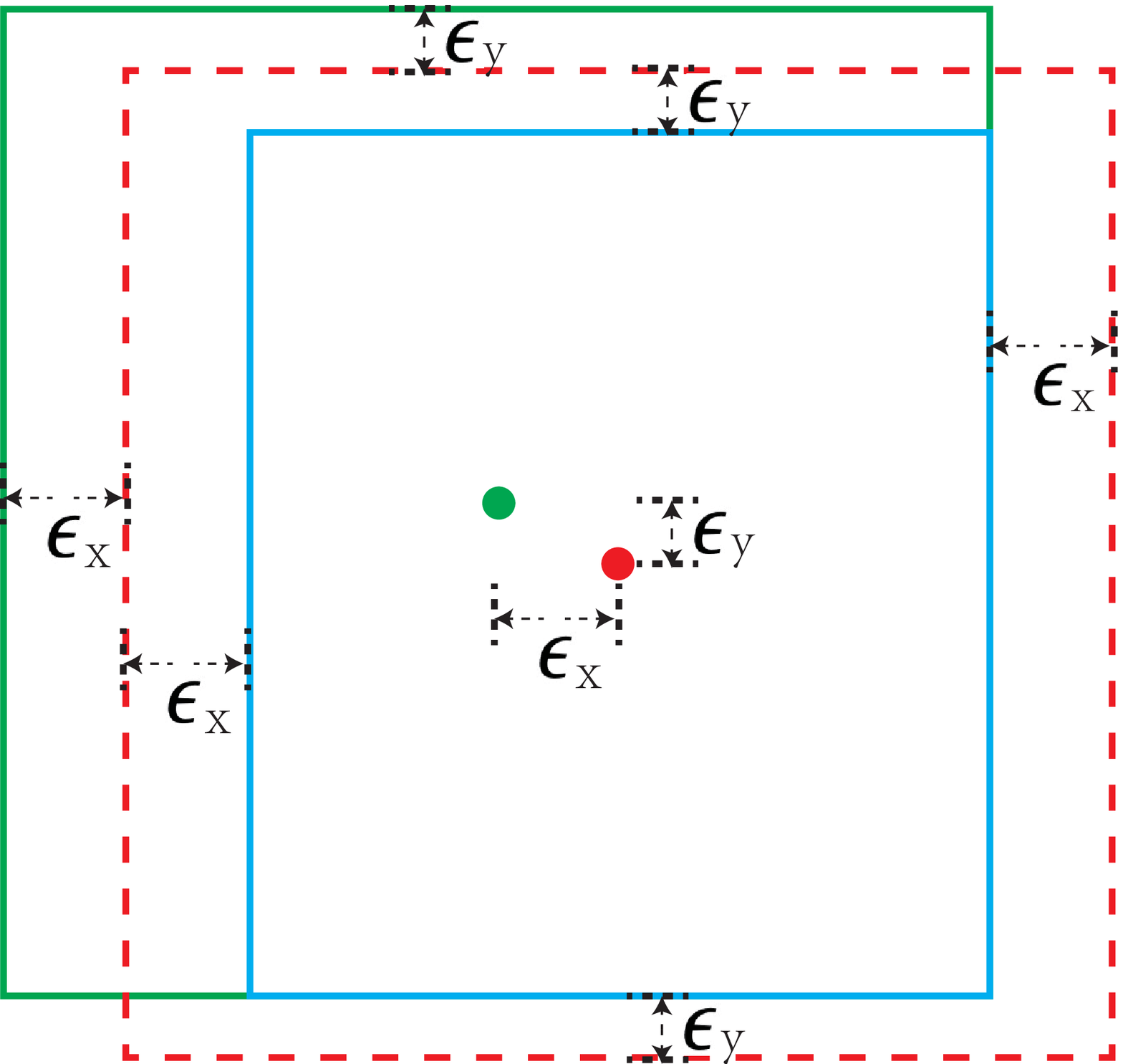}
  \caption{A sample of noise augmentation to a regional proposal box. The green box is the original box, the dashed red box is the result of center movement, and the blue box is the result box after augmentation.}
  \label{fig:noise_proposal}
\end{center}
\end{figure}

It is worth mentioning that adding noise to the regional proposal boxes can be interpreted as a movement of these boxes. Since we set initial boxes to Image Size, any movement $\epsilon$ of the center leads to $2\epsilon$ reduction of height or width, as shown in Figure~\ref{fig:noise_proposal}.

\subsection{Many-to-One Label Assignment}
\label{sec:many_to_one_label_assignment}
As discussed in section~\ref{sec:introduction} and ~\ref{sec:related_work}, label assignment plays a key role in the object detection models. SparseR-CNN employs Hungarian algorithm~\cite{carion2020end} to perform one-to-one label assignment so that the Non Maximum Suppression (NMS) can be removed from the processing pipeline. However, as aforementioned, tables in documents are usually large and have no overlaps, meaning that applying NMS to the TD problem doesn't necessarily lead to some drawbacks caused by the NMS, such as the performance degradation caused by the object overlaps~\cite{bodla2017soft}. Moreover, the cascaded Dynamic Heads take input proposal features and boxes with different qualities, making it necessary to determine the label assignment dynamically. Many studies~\cite{ge2021ota, hong2022dynamic, du20211st} have demonstrated that Many-to-One label assignment can bring benefits to the model performance. Therefore, we adapt SimOTA~\cite{du20211st} as the base label assignment method in this study.

SimOTA is a simplified version of OTA~\cite{ge2021ota}, which can avoid the complex optimization process of OTA. More specifically, SimOTA directly uses the top-k candidates whose centers are in the ground truth bounding boxes as the positive samples, as defined by Equation~\ref{eq:sim_ota_matching_cost}, which contains a classification cost, a regression cost, and a center cost. It is worth mentioning that the cost function of SparseR-CNN is the sum of the cls\_cost and regression\_cost in Equation~\ref{eq:sim_ota_matching_cost}.  

\begin{equation}
\label{eq:sim_ota_matching_cost}
\begin{aligned}
cost_{SimOTA} = \underbrace{\lambda_{cls}\cdot cost_{cls}}_{cls\_cost} + \underbrace{\lambda_{l1}\cdot cost_{l1} + \lambda_{giou}\cdot cost_{giou}}_{regression\_cost} + \underbrace{\lambda_{center} \cdot cost_{center} }_{center\_cost}
\end{aligned}
\end{equation}

SimOTA employs a dynamic method to determine the number of positive samples assigned to each ground truth box using the sum of the top 10 IoU scores between a ground truth box and its corresponding prediction boxes without considering the difference of Dynamic Heads. Considering that the inputs of Dynamic Head $t+1$ should contain higher quality boxes than that of $t$ after the refinement of Dynamic Head $t$, the Dynamic Head $t+1$ should have more positive samples. Therefore, we further extend this dynamic method of SimOTA by adding a scheduling scheme as defined by Equation~\ref{eq:ota_scheduling}, where $N$ is the number of Dynamic Heads, $IoU_i$ is the IoU matrix between the predictions and the $i$th ground truth, $n$ is a hyper parameter and $k_t^i$ means the number of positive samples assigned to the $i$th ground truth for Dynamic Head $t$.

\begin{equation}
\label{eq:ota_scheduling}
\begin{aligned}
 k_t^i = SUM(TOPK(IoU_i, n - 0.5 * (N - t))), t \in [1, N]
\end{aligned}
\end{equation}

At last, we can define the loss function as the sum of all the Dynamic Heads' loss, as defined by Equation~\ref{eq:loss_function}. For the classification loss, we simply use cross entropy and focal loss~\cite{lin2017focal} for binary and multi-class classification, respectively.

\begin{equation}
\label{eq:loss_function}
\begin{aligned}
 \mathcal{L} = \sum_{t=1}^{N} loss_t = \sum_{t=1}^{N} \lambda_{cls} loss_{cls}^t +  \lambda_{l1} loss_{l1}^t + \lambda_{giou} loss_{giou}^t
\end{aligned}
\end{equation}

It is worth mentioning that some studies, such as YOLOX~\cite{ge2021yolox} and Dynamic SparseR-CNN~\cite{hong2022dynamic}, use similar Label Assignment approaches. Dynamic SparseR-CNN introduces an assignment scheduling scheme to the OTA method~\cite{ge2021ota} to dynamically adjust the number of positive label assignments. However, the OTA method requires a complex optimization procedure, which is significantly more time-consuming than SimOTA. Therefore, in this study, we leverage SimOTA and further improve it by adapting the assignment scheduling scheme and the proposed ICS loss function.

\subsection{Information Coverage Score}
\label{sec:information_coverage_score}
\begin{figure}[htp]
\begin{center}
  \includegraphics[width=0.8\columnwidth]{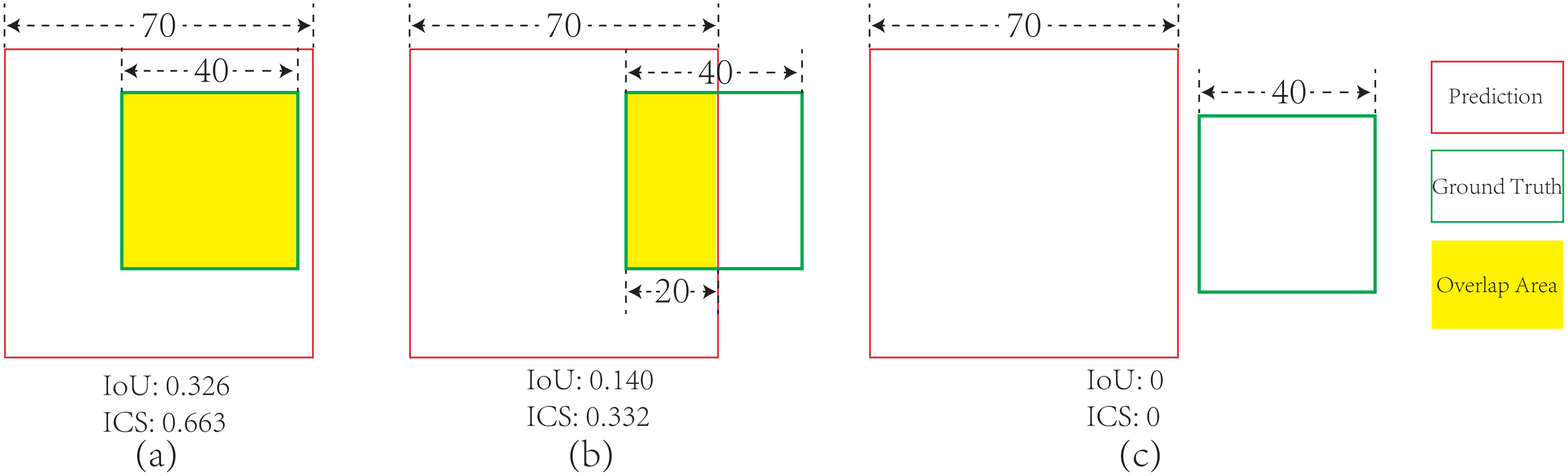}
  \caption{Three cases for the Information Coverage Score. $\lambda$ is set to 0.5. All the boxes are squares.}
  \label{fig:ics}
\end{center}
\end{figure}

As discussed in section~\ref{sec:introduction}, the IoU score cannot directly reflect the information coverage of the prediction boxes. This section discusses our proposed decoupled IoU, the Information Coverage Score (ICS). Assume that $G$ and $P$ are the ground truth and prediction boxes, respectively. IoU is the ratio of the intersection of $G$ and $P$ to the union of $G$ and $P$, as defined by Equation~\ref{eq:iou}. In contrast, ICS contains a ground truth coverage term (GT\_Coverage) and a prediction coverage term (Pred\_Coverage), as defined by Equation~\ref{eq:ics}. The ground truth coverage term is the ratio of the intersection of $G$ and $P$ to the $G$, which can directly measure the information covered by the prediction box. Similarly, the prediction coverage score is defined as the ratio of the intersection of $G$ and $P$ to the $P$. Figure~\ref{fig:ics} shows three cases for the calculation of ICS, in which green boxes, red boxes, and yellow areas represent the ground truth boxes, prediction boxes, and their intersection areas. It is worth mentioning that the proposed ICS can be used to replace IoU in a variety of IoU-based loss functions, such as GIoU loss~\cite{rezatofighi2019generalized} and DIoU loss~\cite{zheng2020distance}. A simple ICS loss can be defined as Equation~\ref{eq:ics_loss}.

\begin{equation}
\label{eq:iou}
\begin{aligned}
IoU = \frac{|G \cap P|}{|G \cup P|}
\end{aligned}
\end{equation}

\begin{equation}
\label{eq:ics}
\begin{aligned}
ICS = \underbrace{\lambda \frac{| G \cap P |}{|G|}}_{GT\_Coverage} + \underbrace{(1-\lambda) \frac{| G \cap P |}{|P|}}_{Pred\_Coverage}
\end{aligned}
\end{equation}

\begin{equation}
\label{eq:ics_loss}
\begin{aligned}
ICS\_loss = 1 - ICS
\end{aligned}
\end{equation}

\section{Experimental Results and Analysis}
\label{sec:experiments}
This section compares the proposed method with state-of-the-art models using IoU-based evaluation metrics and loss functions. Then, we conduct experiments to demonstrate the benefits of using ICS as the evaluation metrics and loss functions. Lastly, an ablation study is conducted to demonstrate the effectiveness of the proposed many-to-one label assignment method and Gaussian Noise Augmented Image Size region proposal.

\subsection{Experiment settings and Main results}
\label{sec:experimental_settings_main_result}
Many datasets have been proposed for the TD problem. We can roughly categorize these datasets into two groups: human-annotated datasets and generated datasets by parsing meta-data. The former dataset type usually has higher-quality annotations, but the number of samples is usually limited. In contrast, the latter type can have a large number of samples but often contain much noise. In this study, we only consider the datasets with high-quality annotations, including ICDAR2017~\cite{gao2017icdar2017}, ICDAR2019~\cite{gao2019icdar}, TNCR~\cite{abdallah2022tncr} and ICT-TD~\cite{xiao2023revisiting} datasets. 

The TD problem is often the pre-processor step of the information extraction tasks, which requires models to avoid missing tables or predicting other document components as tables. Therefore, the widely used evaluation mAP~\cite{lin2014microsoft} for the detection models cannot fulfill this requirement per se. Hence, we use Precision, Recall, and F1 scores as evaluation metrics, which are also widely used in other studies~\cite{gobel2013icdar, gao2017icdar2017, gao2019icdar, abdallah2022tncr, xiao2023revisiting}. However, the IoU thresholds for these metrics vary among studies, making it hard to compare these models directly. Therefore, in this study, we align the evaluation metric for all datasets and use Weighted Average F1, as defined in Equation 8, as evaluation metric whose thresholds are 60\%, 70\%, 80\% and 90\%, and attach detailed results containing Precision, Recall and F1 under the IoU thresholds from 50\% to 95\% with a 5\% interval in  ~\ref{sec:appendix}. A. Notably, the evaluation metric used here is identical to the one used in ICDAR2019 competition~\cite{gao2019icdar}, and other evaluation metrics, such as mAP. The metrics in other competitions~\cite{gobel2013icdar, gao2017icdar2017} can be found in the detailed experimental results in section~\ref{sec:appendix}.

\begin{equation}
\label{eq:weighted_f1_calculation}
\begin{aligned}
\text{Weighted Avg. F1} = \frac{\sum_{i=1}^{4} IoU_i \cdot F1@IoU_i}{\sum_{i=1}^4IoU_i}
\end{aligned}
\end{equation}

\begin{table}[ht!]
\caption{Key training parameters of the proposed model.}
\centering
\begin{tabular}{ l | c | l  }
\hline
\label{table:modek_key_parameters} 
Parameter & Value & Description \\
\hline
IMS\_PER\_BATCH & 16 & number of training samples in an iteration \\
MAX\_ITER & 40,600 &  total number of mini-batch  \\
STEPS & 29,000 & the mini-batch to apply the learning rate schedule  \\
SCHEDULER & MultiStepLR & the scheduler to change the learning rate \\
BASE\_LR & 2.5e-05 &  the learning rate before applying the scheduler \\
WEIGHTS & r50\_300pro\_3x\_model & initialization weight of the model \\
NUM\_HEADS & 6 & the number of Dynamic Head \\
NUM\_PROPOSALS & 300 & number of region proposals\\
OPTIMIZER & AdamW & the optimizer to train the model\\
LABEL\_N & 8 & the hyper parameter $N$ defined by Equation~\ref{eq:ota_scheduling} \\
NOISE\_MEAN & 0 & mean value of the Gaussian Noise\\
NOISE\_VAR & 0.01 & variance value of the Gaussian Noise \\
NMS\_THRESH & 0.9 & non-maximum suppression threshold\\
\hline
\end{tabular}
\end{table}

For the benchmark models, we include all the types of popular object detection models discussed in section~\ref{sec:related_work}, including FasterR-CNN~\cite{ren2015faster}, MaskR-CNN~\cite{he2017mask}, TableDet~\cite{fernandes2022tabledet}, DiffusionDet~\cite{chen2022diffusiondet}, Deformable-DETR~\cite{zhu2020deformable}, SparseR-CNN~\cite{sun2021sparse}, RetinaNet~\cite{lin2017focal}, FCOS~\cite{tian2019fcos}, YOLOX-X~\cite{ge2021yolox}, YOLOR-X~\cite{wang2021you}, YOLOv5-X~\cite{yolov5}, YOLOv7-X~\cite{wang2022yolov7}, YOLOv8-X~\cite{yolov8_ultralytics}. We used the default settings of their implementations, trained FasterR-CNN, MaskR-CNN, TableDet, DiffusionDet, Deformable-DETR, and SparseR-CNN for 120 epochs, and other one-stage detectors for 300 epochs. The detailed settings of these benchmark models are included in section~\ref{sec:appendix}.
Our proposed SparseTableDet is built on the code base of SparseR-CNN, and the key training parameters are summarized in Table~\ref{table:modek_key_parameters}. It is worth mentioning that the parameter names in Table~\ref{table:modek_key_parameters} are aligned with the names in Detectron2~\cite{wu2019detectron2}. More specifically, IMS\_PER\_BATCH is the total number of training samples in an iteration. MAX\_ITER and STEPS refer to the total number of mini-batch used in the training and the mini-batch to apply the learning rate scheduler, respectively. WEIGHTS is the initialization weight of the model. In our implementation, we use the SparseR-CNN model pre-trained with COCO dataset~\cite{lin2014microsoft} as the initialization weight. NUM\_HEADS and LABEL\_N are two custom parameters in the proposed SparseTableDet, which refer to the number of Dynamic Head and the hyper parameter $N$ defined by Equation~\ref{eq:ota_scheduling}, as discussed in section~\ref{sec:proposed_method}. At last, the model is trained with AdamW~\cite{loshchilov2017decoupled} optimizer.

\begin{table*}[t]
\caption{Experimental results on ICDAR2017 dataset.}
\centering
\begin{tabular}{ c c c c c c}
 \toprule
\label{table:icdar2017_results_weighted_F1}
Model & \multicolumn{4}{c}{F1} & Weighted Average F1\\ 
 & IoU(60\%) & IoU(70\%)  & IoU(80\%) & IoU(90\%) & \\ \midrule
RetinaNet & 96.0 & 93.4 & 91.7 & 87.3 & 91.6  \\ 
FCOS & 96.0 & 93.8 & 92.1 & 87.6 & 91.9  \\ 
YOLOX-X & 97.7 & 95.5 & 92.3 & 80.0 & 90.4  \\
YOLOR-X & 97.1 & 95.3 & 93.4 & 89.7 & 93.5  \\ 
YOLOV5-X & 98.5 & 96.8 & 95.4 & 91.9 & 95.3  \\ 
YOLOV7-X & 97.6 & 96.3 & 94.6 & 91.5 & 94.6  \\ 
YOLOV8-X & 97.9 & 96.2 & 95.3 & 92.5 & 95.2  \\ 
FasterR-CNN & 97.1 &  96.0  & 93.8 & 89.6 & 93.7  \\ 
MaskR-CNN & 96.8 &  96.0 & 94.8 & 91.1 & 94.4  \\ 
TableDet & 98.8 &  97.1  &  95.0 & 90.4 & 94.9  \\ 
DiffusionDet & 98.3 & 97.0 & 94.9 & 90.3 & 94.7  \\ 
Deformable-DETR & 97.5 &  96.7  & 94.4 & 91.4 & 94.6  \\ 
SparseR-CNN & 98.3 &  97.9  & 96.1 & 94.0 & 96.3  \\ 
SparseTableDet (Proposed) & $\mathbf{99.5}$ & $\mathbf{99.4}$ & $\mathbf{98.2}$ & $\mathbf{94.8}$ & $\mathbf{97.7}$ \\
\bottomrule
\end{tabular}
\end{table*}

\begin{table*}[t]
\caption{Experimental results on ICDAR2019 dataset.}
\centering
\begin{tabular}{ c c c c c c}
 \toprule
\label{table:icdar2019_results_weighted_F1}
Model & \multicolumn{4}{c}{F1} & Weighted Average F1\\ 
 & IoU(60\%) & IoU(70\%)  & IoU(80\%) & IoU(90\%) & \\ \midrule
RetinaNet & 98.0 & 96.7 & 94.5 & 86.8 & 93.4  \\ 
FCOS & 97.6 & 96.5 & 93.6 & 85.7 & 92.7  \\ 
YOLOX-X & 97.1 & 96.0 & 94.6 & 89.2 & 93.8  \\ 

YOLOR-X & 98.6 & 98.2 & 97.2 & 93.4 & 96.6  \\ 

YOLOV5-X & 99.0 & 98.9 & 98.2 & 95.7 & 97.8  \\ 
YOLOV7-X & 99.2 & 98.6 & 98.0 & 94.1 & 97.2  \\ 
YOLOV8-X & 99.2 & 99.1 & 98.1 & 94.7 & 97.5  \\ 
FasterR-CNN & 97.4 & 96.2 & 95.0 & 90.4 & 94.4  \\ 
MaskR-CNN & 98.2 & 97.0 & 95.8 & 91.9 & 95.4  \\ 
TableDet & 98.1 &  96.8  & 94.9 & 91.5 & 94.9  \\ 
DiffusionDet & 98.9 &  97.4  & 95.8 & 91.3 & 95.5  \\ 
Deformable-DETR & 98.4 &  97.9  & 96.5 & 92.7 & 96.0  \\ 
SparseR-CNN & 98.6 &  98.1  & 97.5 & 94.9 & 97.1  \\ 
SparseTableDet (Proposed) & $\mathbf{99.3}$ & $\mathbf{99.1}$ & $\mathbf{98.9}$ & $\mathbf{96.3}$ & $\mathbf{98.3}$ \\
\bottomrule
\end{tabular}
\end{table*}

\begin{table*}[t]
\caption{Experimental results on TNCR dataset.}
\centering
\begin{tabular}{ c c c c c c}
 \toprule
\label{table:tncr_results_weighted_F1}
Model & \multicolumn{4}{c}{F1} & Weighted Average F1\\ 
 & IoU(60\%) & IoU(70\%)  & IoU(80\%) & IoU(90\%) & \\ \midrule
RetinaNet & 92.7 &  92.0  & 90.6 & 84.8 & 89.6  \\ 
FCOS & 90.8 &  89.9  & 88.8 & 83.3 & 87.8  \\ 
YOLOX-X & 90.6 &  89.3  & 86.1 & 79.6 & 85.8  \\ 
YOLOR-X & 94.2 &  93.4  & 91.8 & 86.4 & 91.0  \\ 
YOLOV5-X & 95.8 & 95.5 & 94. & 89.6 & 93.5  \\
YOLOV7-X & 95.2 & 95.0 & 93.7 & 89.3 & 93.0  \\
YOLOV8-X & 96.1 & 95.5  & 94.6 & 90.1 & 93.7  \\
FasterR-CNN & 91.5 & 91.0 & 90.3 & 84.4 & 88.9  \\ 
MaskR-CNN & 92.5 & 92.2 & 90.9 & 84.7 & 89.6  \\ 
TableDet & 94.7 & 94.4 & 93.3 & 87.7 & 92.2 \\
Deformable-DETR & 94.4 & 94.1 & 92.9 & 89.3 & 92.4 \\
DiffusionDet & 95.4 &  94.6  & 93.1 & 88.5 & 92.5  \\ 
SparseR-CNN & 95.1 & 94.9 & 94.4 & 90.9 & 93.6  \\ 
SparseTableDet (Proposed) & $\mathbf{96.3}$ & $\mathbf{96.2}$ & $\mathbf{95.8}$ & $\mathbf{92.7}$ & $\mathbf{95.1}$\\
\bottomrule
\end{tabular}
\end{table*}

\begin{table*}[t]
\caption{Experimental results on the ICT-TD dataset.}
\centering
\begin{tabular}{ c c c c c c}
 \toprule
\label{table:icttd_results_weighted_F1}
Model & \multicolumn{4}{c}{F1} & Weighted Average F1\\ 
 & IoU(60\%) & IoU(70\%)  & IoU(80\%) & IoU(90\%) & \\ \midrule

RetinaNet & 95.8 & 93.6 & 91.0 & 83.7 & 90.4 \\
FCOS & 91.8 & 90.4 & 87.9 & 82.3 & 87.6 \\
YOLOX-X & 95.8 & 93.6 & 90.1 & 81.6 & 89.5 \\
YOLOR-X & 97.5 & 96.0 & 94.3 & 89.0 & 93.8 \\
YOLOV5-X & 98.0 & 97.2 & 95.8 & 91.7 & 95.3 \\
YOLOV7-X & 98.6 & 97.6 & 95.7 & 92.6 & 95.8 \\
YOLOV8-X & 97.9 & 97.2 & 95.6 & 92.3 & 95.4 \\
FasterR-CNN & 96.8 & 94.7 & 92.9 & 86.8 & 92.3 \\
MaskR-CNN & 96.2 & 94.8 & 92.8 & 87.9 & 92.5 \\
TableDet & 96.9 & 95.7 & 93.6 & 89.1 & 93.4 \\
DiffusionDet & 97.6 & 96.8 & 95.5 & 91.1 & 94.9 \\
Deformable-DETR & 97.4 & 96.5 & 95.0 & 91.2 & 94.7 \\
SparseR-CNN & 97.1 & 95.9 & 94.3 & 90.4 & 94.1 \\
SparseTableDet (Proposed) & $\mathbf{98.2}$ & $\mathbf{97.9}$ & $\mathbf{97.2}$ & $\mathbf{94.2}$ & $\mathbf{96.7}$ \\
\bottomrule
\end{tabular}
\end{table*}

The experimental results for the ICDAR2017, ICDAR2019, TNCR and ICT-TD datasets are shown in Table ~\ref{table:icdar2017_results_weighted_F1}, ~\ref{table:icdar2019_results_weighted_F1}, ~\ref{table:tncr_results_weighted_F1} and ~\ref{table:icttd_results_weighted_F1}, respectively. The experimental results show that the proposed SparseTableDet can consistently outperform the state-of-the-art benchmark models regarding the Weighted Average F1 score. We also compare our proposed model with other state-of-the-art models optimized for the TD problem following the evaluation protocols of ICDAR2013, ICDAR2017, and ICT-TD datasets, and include the results in section~\ref{sec:appendix}. With these competition evaluation protocols, our proposed method can still consistently outperform the benchmark models.

\subsection{ICS for model training and evaluation}
\label{sec:ics_for_model_evaluation}
\begin{figure}[htp]
\begin{center}
  \includegraphics[width=0.8\columnwidth]{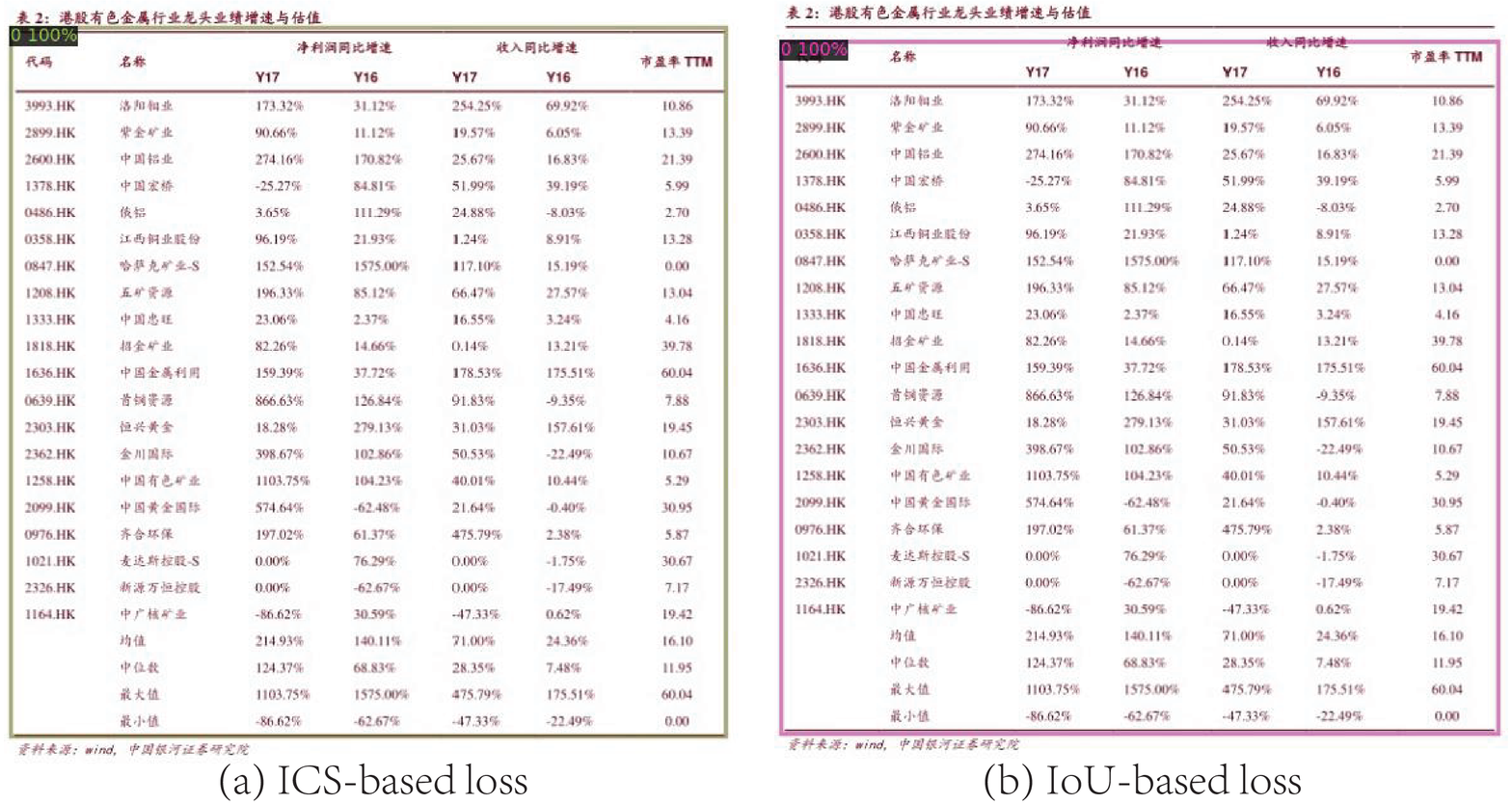}
  \caption{Prediction samples of models trained with ICS-based loss and IoU-based loss.}
  \label{fig:prediction_results1}
\end{center}
\end{figure}

As discussed in section~\ref{sec:information_coverage_score}, GT\_Coverage term in the ICS is a direct metric to measure whether the prediction box covers all the target content. This section uses the GT\_Coverage as the evaluation metric to evaluate the model performance. More specifically, we replace the IoU score defined in Equation~\ref{eq:weighted_f1_calculation} with GT\_Coverage to define a new Weighted Average F1 score as the evaluation metric, as defined by Equation~\ref{eq:weighted_f1_calculation_gts}. To demonstrate the effectiveness of ICS as the loss function, we replace the cost\_{giou} in Equation~\ref{eq:sim_ota_matching_cost} and GIoU loss in Equation~\ref{eq:loss_function} with their ICS-based counterparts. For simplicity, we use $M_{giou}$ and $M_{ics}$ to represent the model trained with GIoU loss and ICS loss, respectively. As shown in Table~\ref{table:ics_and_ics_loss}, when the IoU-based metrics are used, $M_{giou}$ can perform better than $M_{ics}$. However, as aforementioned in section ~\ref{sec:introduction} and ~\ref{sec:information_coverage_score}, GT\_Coverage term defined in the ICS is a direct measure to evaluate the ground truth information covered by the prediction. GT\_Coverage-based evaluation metric is used, $M_{ics}$ can perform better. Figure~\ref{fig:prediction_results1} shows two prediction results of $M_{giou}$ and $M_{ics}$. Using an ICS-based loss function can encourage the model to alleviate the information loss during the optimization process because the GT\_Coverage term in the ICS is a direct measure of the information loss and is more sensitive to the information loss. We include some other prediction samples of these two models in ~\ref{sec:appendix}. Notably, bias might be introduced when using the GT Coverage score as the evaluation metric because the GT Coverage score cannot reflect the difference of prediction boxes once the predictions can cover the ground truth. However, the proposed ICS and GT Coverage scores can provide more insights regarding the quality of predictions and complement IoU-based metrics.

\begin{equation}
\label{eq:weighted_f1_calculation_gts}
\begin{aligned}
\text{Weighted Avg. F1} = \frac{\sum_{i=1}^{4} GT\_Coverage_i \cdot F1@GT\_Coverage_i}{\sum_{i=1}^4GT\_Coverage_i}
\end{aligned}
\end{equation}

\begin{table*}[t]
\caption{Experimental results on ICDAR2019 dataset evaluated by Weighted Average F1 scores using GT\_Coverage and IoU as thresholds.}
\centering
\begin{tabular}{ c c c c c c c}
 \toprule
\label{table:ics_and_ics_loss}
Loss Function & Metric & \multicolumn{4}{c}{F1} & Weighted Average F1\\ 
 & & 60\% & 70\%  & 80\% & 90\% & \\ \midrule
GIoU & IoU & 99.3 & 99.1 & 98.9 & 96.3 & 98.3 \\
ICS  & IoU & 99.4 & 98.8 & 98.1 & 92.9 & 97.0 \\
\hline
GIoU & GT\_C & 99.6 & 99.5 & 99.1 & 98.3 & 99.1 \\
ICS  & GT\_C & 99.7 & 99.6 & 99.5 & 98.6 & 99.3 \\
\bottomrule
\end{tabular}
\end{table*}

\subsection{Ablation Study}
\label{sec:ablation_study}
This section discusses the effectiveness of the proposed Image Size regional proposals, Noise Augmented Proposals, and Many-to-One label assignment method. We use SparseR-CNN as the baseline model in this section, which uses Hungarian Matching for the label assignment and random region proposals. We use the ICDAR2019 dataset to conduct experiments and use the Weighted Average F1 scores defined by Equation~\ref{eq:weighted_f1_calculation} and Equation~\ref{eq:weighted_f1_calculation_gts} as the evaluation metrics. It is worth mentioning that we choose 60\%, 70\%, 80\% and 90\% as thresholds to align with the metric in section~\ref{sec:experimental_settings_main_result}. The experimental results are shown in Table~\ref{table:ablation_study_on_icdar2019} and ~\ref{table:ablation_study_on_icdar2019_ics}, where SparseR-CNN(R), SparseR-CNN(I) are the SparseR-CNN initialized with the random proposals and image size proposals, respectively. ManytoOne and Noise represent the proposed many-to-one label assignment and the Noise Augmentation to regional proposals. The experimental results show that using image size region proposals, adding noise to the regional proposals, and Many-to-One label assignment can improve the performance of the base SparseR-CNN model.

\begin{table*}[t]
\caption{The effectiveness of each component using IoU scores as thresholds.}
\centering
\begin{tabular}{c c c c c c c c}
 \toprule
\label{table:ablation_study_on_icdar2019}
 & \multicolumn{4}{c}{F1} & Weighted Average F1\\ 
 & IoU(60\%) & IoU(70\%)  & IoU(80\%) & IoU(90\%) & \\ \midrule
SparseR-CNN (R) & 98.6 & 98.1 & 97.5 & 94.9 & 97.1 \\
SparseR-CNN (I) & 99.1 & 98.7 & 98.2 & 95.3 & 97.6 \\ 
I+ManytoOne & 99.4 & 99.1 & 98.8 & 95.3 & 97.9 \\ 
I+Noise+ManytoOne & 99.3 & 99.1 & 98.9 & 96.3 & 98.3 \\
\bottomrule
\end{tabular}
\end{table*}

\begin{table*}[t]
\caption{The effectiveness of each component using GT Coverage scores as thresholds.}
\centering
\begin{tabular}{c c c c c c c c}
 \toprule
\label{table:ablation_study_on_icdar2019_ics}
 & \multicolumn{4}{c}{F1} & Weighted Average F1\\ 
 & GT\_C (60\%) & GT\_C(70\%)  & GT\_C(80\%) & GT\_C(90\%) & \\ \midrule
SparseR-CNN (R) & 98.7 & 98.7 & 97.9 & 96.7 & 97.9 \\
SparseR-CNN (I) & 99.1 & 99.0 & 98.3 & 97.6 & 98.4 \\ 
I+ManytoOne & 99.4 & 99.4 & 99.2 & 98.1 & 98.9 \\ 
I+Noise+ManytoOne & 99.6 & 99.5 & 99.1 & 98.3 & 99.1 \\
\bottomrule
\end{tabular}
\end{table*}

\section{Conclusion}
\label{sec:conclusion}
In this study, we propose to use SparseR-CNN~\cite{sun2021sparse} as the base model and further improve the model by introducing Noise Augmented region proposal generation, Many-to-One label assignment, and a decoupled IoU. The experimental results show that the proposed method can consistently outperform benchmark models regarding the Weighted Average F1 score on various datasets. Furthermore, considering the requirement of TD applications, we propose to use GT\_Coverage in ICS to replace IoU to act as the evaluation metric and use ICS to replace IoU to derive ICS-based loss functions. The experimental results demonstrate that the GT\_Coverage can be a better metric reflecting the prediction's information loss, and ICS-based loss can guide models to cover more information of the target objects. In this study, we assume that all the area in a ground truth box contains information without considering the inner structure of tables. However, some tables contain extra spaces, meaning that some smaller prediction boxes than their ground truth boxes do not lead to any information loss. Therefore, it can be a direction to consider the inner structure of a table to build more reliable evaluation metrics for the TD applications. Besides, as aforementioned in section~\ref{sec:ics_for_model_evaluation}, the proposed GT\_Coverage score cannot reflect the difference of box size once the prediction box can cover the whole ground truth. Therefore, it can be another direction to integrate the size of boxes to the GT\_Coverage score to make it more versatile.



 \bibliographystyle{elsarticle-num} 
 \bibliography{reference.bib}

\appendix

\section{Appendix}
\label{sec:appendix}
\subsection{Model implementations and settings}
In this section, we list the implementations and configuration files of the baseline models, including RetinaNet~\cite{lin2017focal}, FCOS~\cite{tian2019fcos}, YOLOX-X~\cite{ge2021yolox}, YOLOR-X~\cite{wang2021you}, YOLOv5-X~\cite{yolov5}, YOLOv7-X~\cite{wang2022yolov7}, YOLOv8-X~\cite{yolov8_ultralytics}, FasterR-CNN~\cite{ren2015faster}, DiffusionDet~\cite{chen2022diffusiondet}, Deformable-DETR~\cite{zhu2020deformable}, and SparseR-CNN~\cite{sun2021sparse}, as summarized in Table~\ref{table:implementations_settings}. It is worth mentioning that we modified the training epochs of the listed configuration files, trained FasterR-CNN, MaskR-CNN, DiffusionDet, Deformable-DETR, and SparseR-CNN for 120 epochs, and trained other one-stage detectors for 300 epochs.

\begin{tabularx}{\textwidth}{c c X}
  \caption{Summary of model implementations and settings}\label{table:implementations_settings}\\ \toprule
   Model & Implementation & Setting File \\  \midrule
RetinaNet & Detectron2 & \url{https://github.com/facebookresearch/detectron2/blob/main/configs/COCO-Detection/retinanet_R_50_FPN_3x.yaml} \\ \hline
FCOS & Detectron2 & \url{https://github.com/facebookresearch/detectron2/blob/main/configs/COCO-Detection/fcos_R_50_FPN_1x.py} \\ \hline
YOLOX & Official codebase & \url{https://github.com/Megvii-BaseDetection/YOLOX/blob/main/exps/default/yolox_x.py} \\ \hline
YOLOR & Official codebase & \url{https://github.com/WongKinYiu/yolor/blob/main/cfg/yolor_csp_x.cfg} \\ \hline
YOLOv5 & Official codebase & \url{https://github.com/ultralytics/ultralytics/blob/main/ultralytics/cfg/models/v5/yolov5.yaml} \\ \hline
YOLOv7 & Official codebase & \url{https://github.com/WongKinYiu/yolov7/blob/main/cfg/training/yolov7x.yaml} \\ \hline
YOLOv8 & Official codebase& \url{https://github.com/ultralytics/ultralytics/blob/main/ultralytics/cfg/models/v8/yolov8.yaml} \\ \hline
FasterR-CNN & Detectron2 & \url{https://github.com/facebookresearch/detectron2/blob/main/configs/COCO-Detection/faster_rcnn_R_50_FPN_3x.yaml} \\ \hline
MaskR-CNN & Detectron2& \url{https://github.com/facebookresearch/detectron2/blob/main/configs/COCO-InstanceSegmentation/mask_rcnn_R_50_FPN_3x.yaml} \\ \hline
DiffusionDet & Official codebase& \url{https://github.com/ShoufaChen/DiffusionDet/blob/main/configs/diffdet.coco.res50.300boxes.yaml} \\ \hline
Deformable-DETR & detrex & \url{https://github.com/IDEA-Research/detrex/blob/main/projects/deformable_detr/configs/deformable_detr_r50_two_stage_50ep.py} \\ \hline
SparseR-CNN & Official codebase& \url{https://github.com/PeizeSun/SparseR-CNN/blob/main/projects/SparseRCNN/configs/sparsercnn.res50.300pro.3x.yaml} \\ \hline
\bottomrule
\end{tabularx}

\subsection{Compared with other Table Detection models}
In this section, we include the experimental results using the evaluation protocols in ICDAR2013, ICDAR2017, and ICT-TD datasets. More specifically, for the ICDAR2013 dataset, the F1 score thresholded by 50\% is the competition evaluation metric. For the ICDAR2017 dataset, Precision, Recall, and  F1 scores thresholded by 60\% and 80\% are used as evaluation metrics. For the ICT-TD dataset, Weighted Average F1 score, as defined in Equation~\ref{eq:weighted_f1_calculation}, is used as the evaluation metric whose thresholds are 80\%, 85\%, 90\%, and 95\%. The experimental results of them are shown in Table~\ref{table:icdar2013_results},~\ref{table:icdar2017_results} and ~\ref{table:icttd_results}. It is worth mentioning that the experimental results of TableDet~\cite{fernandes2022tabledet}, DeCNT~\cite{siddiqui2018decnt}, YOLOv3-TD~\cite{huang2019yolo}, DeepDeSRT~\cite{schreiber2017deepdesrt}, TableNet~\cite{paliwal2019tablenet}, GAN-TD~\cite{li2019gan}, in Table~\ref{table:icdar2013_results},~\ref{table:icdar2017_results} are from study~\cite{fernandes2022tabledet}.

\begin{table*}[t]
\caption{Experimental results on ICDAR2013 dataset (IoU = 50\%).}
\centering
\begin{tabular}{ c c c c }
 \toprule
\label{table:icdar2013_results}
Model & Precision & Recall  & F1\\ \midrule
CascadeTabNet~\cite{prasad2020cascadetabnet} & 100 &  100  & 100  \\
TableDet~\cite{fernandes2022tabledet} & 100 &  100  & 100  \\  
DeCNT~\cite{siddiqui2018decnt} & 99.6 &  99.6  & 99.6  \\ 
YOLOv3-TD~\cite{huang2019yolo} & 94.9 &  100  & 97.3  \\ 
DeepDeSRT~\cite{schreiber2017deepdesrt} & 97.4 & 96.2 & 96.8\\
TableNet~\cite{paliwal2019tablenet} & 97.0 & 96.3 & 96.6 \\
SparseTableDet (Proposed) & 100 & 100 & 100 \\
\bottomrule 
\end{tabular}
\end{table*}

\begin{table*}[t]
\caption{Experimental results on ICDAR2017 dataset.}
\centering
\begin{tabular}{ c c c c c }
 \toprule
\label{table:icdar2017_results}
IoU Threshold & Model & Precision & Recall  & F1\\ \midrule

 & TableDet~\cite{fernandes2022tabledet} & 98.8 &  99.7  & 99.3  \\  
& YOLOv3-TD~\cite{huang2019yolo} & 97.2 &  97.8  & 97.5  \\ 
60\%& DeCNT~\cite{siddiqui2018decnt} & 96.5 &  97.1  & 96.8  \\ 
& GAN-TD~\cite{li2019gan} & 94.4 & 94.4 & 94.4 \\
& SparseTableDet (Proposed) & $\mathbf{99.1}$ & $\mathbf{100.0}$ & $\mathbf{99.5}$ \\
\hline
 & TableDet~\cite{fernandes2022tabledet} & $\mathbf{97.4}$ &  98.4  & 97.9  \\  
& YOLOv3-TD~\cite{huang2019yolo} & 96.8 &  97.5  & 97.1  \\ 
80\%& DeCNT~\cite{siddiqui2018decnt} & 96.7 &  93.7  & 95.2  \\ 
& GAN-TD~\cite{li2019gan} & 90.3 & 90.3 &  90.3 \\
& SparseTableDet (Proposed) & 96.7 & $\mathbf{99.7}$ & $\mathbf{98.2}$ \\
 \bottomrule 
\end{tabular}
\end{table*}

\begin{table*}[t]
\caption{Experimental results on the ICT-TD dataset.}
\centering
\begin{tabular}{ c c c c c c}
 \toprule
\label{table:icttd_results}
Model & \multicolumn{4}{c}{F1} & Weighted Average F1\\ 
 & IoU(80\%) & IoU(85\%)  & IoU(90\%) & IoU(95\%) & \\ \midrule
TableDet~\cite{xiao2023revisiting} & 93.6 & 91.6 & 89.1 & 75.7 & 87.1  \\ 
DiffusionDet~\cite{xiao2023revisiting} & 95.5 & 94.2 & 91.1 & 76.4 & 88.9 \\
Deformable-DETR~\cite{xiao2023revisiting} & 95.0 & 93.9 & 91.2 & $\mathbf{83.0}$ & 90.5 \\
SparseR-CNN~\cite{xiao2023revisiting} & 94.3 & 93.0 & 90.4 & 78.8 & 88.8 \\
SparseTableDet (Proposed) & $\mathbf{97.2}$ & $\mathbf{96.4}$ & $\mathbf{94.2}$ & 81.8 & $\mathbf{92.1}$ \\
\bottomrule
\end{tabular}
\end{table*}

\subsection{Detailed experimental results}
In this appendix section, we include the detailed experimental results on the TNCR and ICT-TD datasets, as shown in Table~\ref{table:detailed_tncr_results} and Table~\ref{table:icttd_detailed_results}. Besides, we also include some prediction results for the models trained with IoU-based and ICS-based losses, as discussed in section~\ref{sec:ics_for_model_evaluation}.

\begin{figure}[htp]
\begin{center}
  \includegraphics[width=1.0\columnwidth]{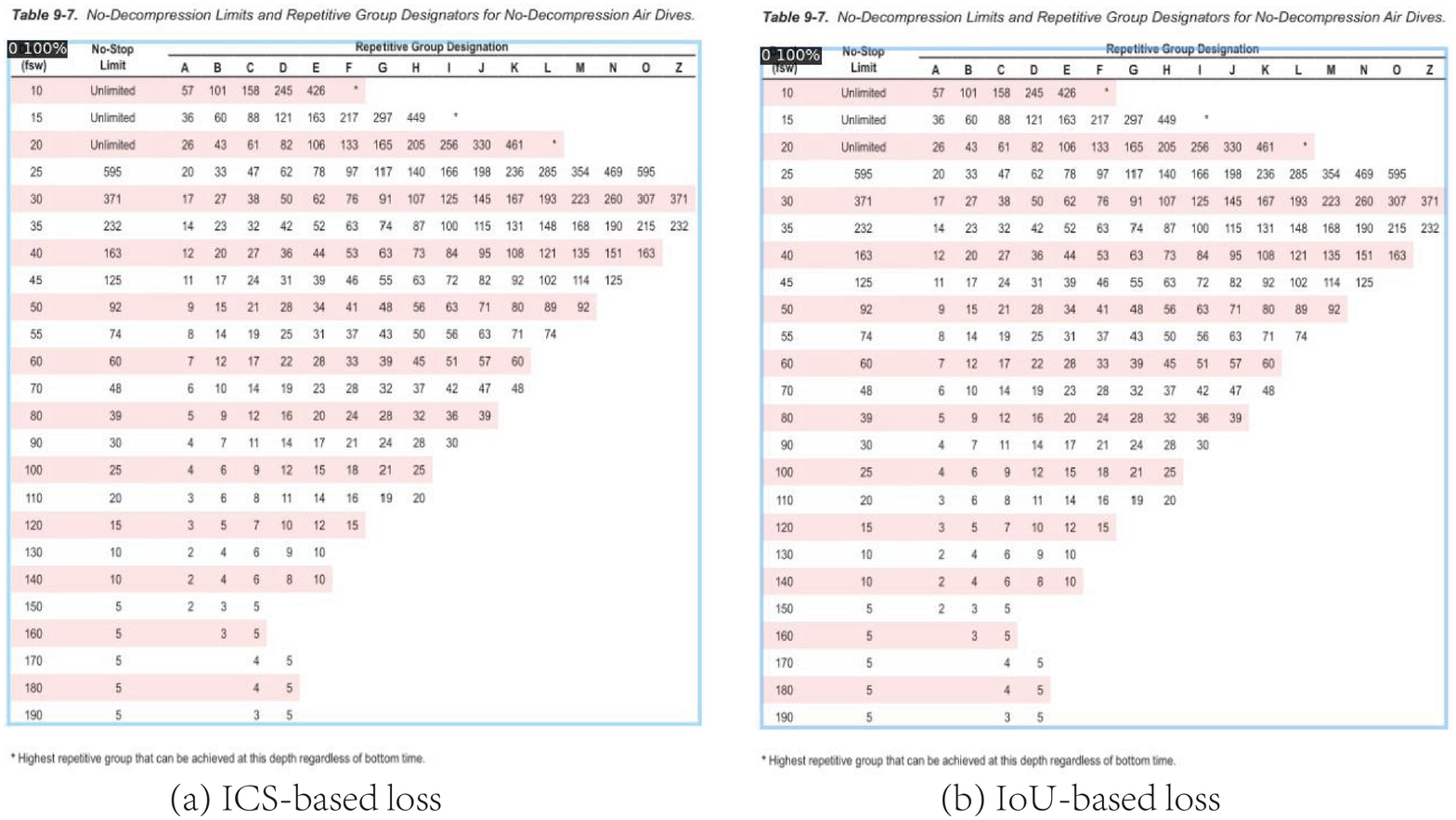}
  \caption{Prediction samples of models trained with ICS-based loss and IoU-based loss.}
  \label{fig:prediction_results2}
\end{center}
\end{figure}

\begin{figure}[htp]
\begin{center}
  \includegraphics[width=1.0\columnwidth]{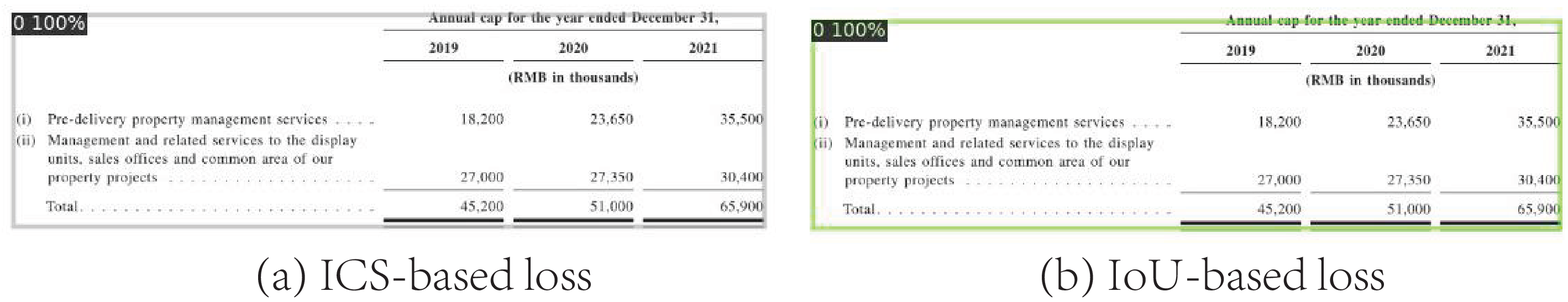}
  \caption{Prediction samples of models trained with ICS-based loss and IoU-based loss.}
  \label{fig:prediction_results3}
\end{center}
\end{figure}

\begin{longtable}[c]{ c c c c c c c c c c c c c c}
\caption{Detailed Experimental results on the ICDAR2017 dataset.} \label{table:detailed_icdar2017_results} \\

\hline 
\multirow{2}{*}{Method} & \multirow{2}{*}{Metric} & \multicolumn{11}{{c}}{IoU}  \\ \cmidrule{3-13}
& &50\% &55\% &60\% &65\% &70\% &75\% &80\% &85\% &90\% &95\% & 50\%:95\% \\ \midrule
\endfirsthead

\multicolumn{13}{c}%
{{\tablename\ \thetable{} Detailed Experimental results on the ICDAR2017 dataset (continued from previous page).}} \\
\hline 
\multirow{2}{*}{Method} & \multirow{2}{*}{Metric} & \multicolumn{11}{{c}}{IoU}  \\ \cmidrule{3-13} 
& &50\% &55\% &60\% &65\% &70\% &75\% &80\% &85\% &90\% &95\% & 50\%:95\% \\ \midrule
\endhead

\hline \multicolumn{13}{r}{{Continued on next page}} \\ \hline
\endfoot
\hline

\endlastfoot
 RetinaNet  & Precision & 96.4 & 96.4 & 95.2 & 94.9 & 92.1 & 90.7 & 90.3 & 88.1 & 85.5 & 75.0 & 90.4\\ 
 &  Recall & 97.8 & 97.8 & 96.9 & 96.3 & 94.7 & 93.8 & 93.2 & 91.6 & 89.1 & 80.1 & 93.1 \\ 
&  F1 & 97.1 & 97.1 & 96.0 & 95.6 & 93.4 & 92.2 & 91.7 & 89.8 & 87.3 & 77.5 & 91.7 \\ 
 \hline
FCOS & Precision & 97.3 & 96.3 & 95.2 & 94.9 & 92.4 & 90.8 & 90.2 & 87.6 & 84.9 & 72.7 & 90.2\\ 
  &  Recall & 98.1 & 97.5 & 96.9 & 96.6 & 95.3 & 94.7 & 94.1 & 92.8 & 90.3 & 82.6 & 93.9 \\ 
 &  F1 & 97.7 & 96.9 & 96.0 & 95.7 & 93.8 & 92.7 & 92.1 & 90.1 & 87.5 & 77.3 & 92.0 \\ 
 \hline
 YOLOX-X & Precision & 97.0 & 97.0 & 96.3 & 95.5 & 94.1 & 93.3 & 91.2 & 87.1 & 78.8 & 54.2 & 88.5\\ 
  &  Recall & 99.4 & 99.4 & 99.1 & 98.4 & 96.9 & 95.6 & 93.5 & 89.4 & 81.3 & 59.5 & 91.3 \\ 
 &  F1 & 98.2 & 98.2 & 97.7 & 96.9 & 95.5 & 94.4 & 92.3 & 88.2 & 80.0 & 56.7 & 89.9 \\ 
 \hline

YOLOR-X & Precision & 97.5 & 97.4 & 96.4 & 95.1 & 94.3 & 93.4 & 92.2 & 92.2 & 88.1 & 79.9 & 92.6\\ 
 &  Recall & 98.8 & 98.8 & 97.8 & 96.9 & 96.3 & 95.3 & 94.7 & 94.4 & 91.3 & 84.1 & 94.8 \\ 
 &  F1 & 98.1 & 98.1 & 97.1 & 96.0 & 95.3 & 94.3 & 93.4 & 93.3 & 89.7 & 81.9 & 93.7 \\ 
 \hline

 YOLOV5-X & Precision & 98.3 & 98.3 & 97.9 & 97.9 & 95.8 & 95.4 & 94.2 & 93.8 & 90.2 & 82.8 & 94.4\\ 
  &  Recall & 99.7 & 99.7 & 99.1 & 99.1 & 97.8 & 97.2 & 96.6 & 96.3 & 93.8 & 86.6 & 96.6 \\ 
 &  F1 & 99.0 & 99.0 & 98.5 & 98.5 & 96.8 & 96.3 & 95.4 & 95.0 & 91.9 & 84.7 & 95.5 \\ 
 \hline

YOLOV7-X & Precision & 98.0 & 97.1 & 97.0 & 95.9 & 95.3 & 94.1 & 93.3 & 93.3 & 89.8 & 80.7 & 93.5\\ 
  &  Recall & 99.1 & 98.4 & 98.1 & 97.5 & 97.2 & 96.6 & 96.0 & 95.6 & 93.2 & 85.1 & 95.7 \\ 
 &  F1 & 98.5 & 97.7 & 97.5 & 96.7 & 96.2 & 95.3 & 94.6 & 94.4 & 91.5 & 82.8 & 94.6 \\ 
 \hline

 YOLOV8-X & Precision & 99.0 & 97.9 & 97.1 & 96.5 & 95.2 & 94.6 & 94.1 & 93.9 & 91.0 & 80.6 & 94.0\\ 
  &  Recall & 100.0 & 99.1 & 98.8 & 98.1 & 97.2 & 96.9 & 96.6 & 96.3 & 94.1 & 85.1 & 96.2 \\ 
 &  F1 & 99.5 & 98.5 & 97.9 & 97.3 & 96.2 & 95.7 & 95.3 & 95.1 & 92.5 & 82.8 & 95.1 \\ 
 \hline
 
FasterR-CNN & Precision & 97.8 & 96.9 & 96.7 & 96.6 & 95.4 & 94.0 & 92.9 & 91.3 & 88.2 & 79.4 & 92.9\\ 
  &  Recall & 98.1 & 97.8 & 97.5 & 97.2 & 96.6 & 95.3 & 94.7 & 93.8 & 91.0 & 83.2 & 94.5 \\ 
 &  F1 & 97.9 & 97.3 & 97.1 & 96.9 & 96.0 & 94.6 & 93.8 & 92.5 & 89.6 & 81.3 & 93.7 \\ 
 \hline

MaskR-CNN & Precision & 97.5 & 96.5 & 96.2 & 96.2 & 95.1 & 94.0 & 93.9 & 92.6 & 89.8 & 68.5 & 92.0\\ 
  &  Recall & 98.1 & 97.8 & 97.5 & 97.2 & 96.9 & 96.0 & 95.6 & 94.7 & 92.5 & 76.6 & 94.3 \\ 
 &  F1 & 97.8 & 97.1 & 96.8 & 96.7 & 96.0 & 95.0 & 94.7 & 93.6 & 91.1 & 72.3 & 93.1 \\ 
 \hline

TableDet & Precision & 99.4 & 98.6 & 98.5 & 98.5 & 96.4 & 95.1 & 94.1 & 93.5 & 89.0 & 78.8 & 94.2\\ 
  &  Recall & 100.0 & 99.7 & 99.1 & 99.1 & 97.8 & 96.6 & 96.0 & 95.0 & 91.9 & 84.1 & 95.9 \\ 
 &  F1 & 99.7 & 99.1 & 98.8 & 98.8 & 97.1 & 95.8 & 95.0 & 94.2 & 90.4 & 81.4 & 95.0 \\ 
 \hline

DiffusionDet & Precision & 98.1 & 98.0 & 97.6 & 97.1 & 96.1 & 94.6 & 93.9 & 92.5 & 89.1 & 77.8 & 93.5\\ 
  &  Recall & 99.7 & 99.7 & 99.1 & 98.8 & 97.8 & 96.3 & 96.0 & 94.4 & 91.6 & 83.5 & 95.7 \\ 
 &  F1 & 98.9 & 98.8  & 98.3 & 97.9 & 96.9 & 95.4 & 94.9 & 93.4 & 90.3 & 80.5 & 94.6 \\ 
 \hline
 
Deformable- & Precision & 97.2 & 97.2 & 96.8 & 96.8 & 95.8 & 93.7 & 92.6 & 91.3 & 89.7 & 78.6 & 93.0\\ 
 DETR &  Recall & 98.8 & 98.4 & 98.1 & 98.1 & 97.5 & 96.6 & 96.3 & 95.0 & 93.2 & 83.2 & 95.5 \\ 
 &  F1 & 98.0 & 97.8 & 97.4 & 97.4 & 96.6 & 95.1 & 94.4 & 93.1 & 87.1 & 80.8 & 94.2 \\ 
 \hline

 SparseR-CNN & Precision & 99.3 & 98.3 & 97.8 & 97.8 & 97.0 & 95.9 & 94.8 & 94.8 & 92.7 & 84.6 & 95.3\\ 
   &  Recall & 100.0 & 99.7 & 99.1 & 99.1 & 98.8 & 98.1 & 97.5 & 97.2 & 95.3 & 89.1 & 97.4 \\ 
 &  F1 & 99.6 & 99.0 & 98.4 & 98.4 & 97.9 & 97.0 & 96.1 & 96.0 & 94.0 & 86.8 & 96.3 \\ 
 \hline
SparseTableDet & Precision & 99.7 & 99.7 & 99.1 & 99.1 & 98.9 & 97.5 & 96.7 & 95.6 & 93.1 & 83.6 & 96.3 \\ 
(Proposed) &  Recall & 100.0 & 100.0 & 100.0 & 100.0 & 100.0 & 100.0 & 99.7 & 99.1 & 96.6 & 89.1 & 98.4 \\ 
 &  F1 & 99.8 & 99.8 & 99.5 & 99.5 & 99.4 & 98.7 & 98.2 & 97.3 & 94.8 & 86.3 & 97.3 \\ 
\end{longtable}

\begin{longtable}[c]{ c c c c c c c c c c c c c c}
\caption{Detailed Experimental results on the ICDAR2019 dataset.} \label{table:detailed_icdar2019_results} \\

\hline 
\multirow{2}{*}{Method} & \multirow{2}{*}{Metric} & \multicolumn{11}{{c}}{IoU}  \\ \cmidrule{3-13}
& &50\% &55\% &60\% &65\% &70\% &75\% &80\% &85\% &90\% &95\% & 50\%:95\% \\ \midrule
\endfirsthead

\multicolumn{13}{c}%
{{\tablename\ \thetable{} Detailed Experimental results on the ICDAR2019 dataset (continued from previous page).}} \\
\hline 
\multirow{2}{*}{Method} & \multirow{2}{*}{Metric} & \multicolumn{11}{{c}}{IoU}  \\ \cmidrule{3-13} 
& &50\% &55\% &60\% &65\% &70\% &75\% &80\% &85\% &90\% &95\% & 50\%:95\% \\ \midrule
\endhead

\hline \multicolumn{13}{r}{{Continued on next page}} \\ \hline
\endfoot

\hline
\endlastfoot
RetinaNet  & Precision & 98.6 & 98.4 & 97.4 & 97.2 & 96.2 & 94.3 & 93.9 & 90.5 & 85.3 & 72.8 & 92.5\\ 
 &  Recall & 99.6 & 99.1 & 98.7 & 98.2 & 97.3 & 96.0 & 95.1 & 92.7 & 88.4 & 78.6 & 94.4 \\ 
&  F1 & 99.1 & 98.7 & 98.0 & 97.7 & 96.7 & 95.1 & 94.5 & 91.6 & 86.8 & 75.6 & 93.4 \\ 
 \hline
FCOS  & Precision & 97.1 & 96.7 & 96.7 & 95.7 & 95.5 & 94.4 & 92.2 & 90.2 & 83.1 & 63.7 & 90.5\\ 
 &  Recall & 98.9 & 98.4 & 98.4 & 97.8 & 97.6 & 96.9 & 95.1 & 93.5 & 88.4 & 75.7 & 94.1 \\ 
&  F1 & 98.0 & 97.5 & 97.5 & 96.7 & 96.5 & 95.6 & 93.6 & 91.8 & 85.7 & 69.2 & 92.3 \\ 
 \hline
YOLOX-X  & Precision & 97.5 & 97.1 & 96.3 & 95.7 & 95.3 & 94.3 & 94.1 & 92.2 & 88.3 & 70.4 & 92.1\\ 
 &  Recall & 98.7 & 98.4 & 98.0 & 97.1 & 96.7 & 96.0 & 95.1 & 93.3 & 90.2 & 74.4 & 93.8 \\ 
&  F1 & 98.1 & 97.7 & 97.1 & 96.4 & 96.0 & 95.1 & 94.6 & 92.7 & 89.2 & 72.3 & 92.9 \\ 
 \hline

YOLOR-X & Precision & 98.7 & 98.7 & 98.2 & 98.2 & 97.5 & 97.5 & 96.6 & 95.6 & 92.5 & 82.1 & 95.5\\ 
&  Recall & 99.6 & 99.6 & 99.1 & 99.1 & 98.9 & 98.7 & 97.8 & 96.4 & 94.2 & 85.3 & 96.9 \\ 
 &  F1 & 99.1 & 99.1 & 98.6 & 98.6 & 98.2 & 98.1 & 97.2 & 96.0 & 93.3 & 83.7 & 96.2 \\ 
 \hline

YOLOV5-X  & Precision & 98.7 & 98.6 & 98.5 & 98.5 & 98.5 & 98.4 & 97.5 & 96.5 & 95.1 & 83.7 & 96.4\\ 
 &  Recall & 99.8 & 99.8 & 99.6 & 99.3 & 99.3 & 99.3 & 98.9 & 97.8 & 96.4 & 86.6 & 97.7 \\ 
&  F1 & 99.2 & 99.2 & 99.0 & 98.9 & 98.9 & 98.8 & 98.2 & 97.1 & 95.7 & 85.1 & 97.0 \\ 
 \hline

YOLOV7-X  & Precision & 99.5 & 98.6 & 98.6 & 98.4 & 98.1 & 97.7 & 97.6 & 96.7 & 93.3 & 81.0 & 95.9\\ 
 &  Recall & 100.0 & 99.8 & 99.8 & 99.1 & 99.1 & 98.7 & 98.4 & 97.8 & 94.9 & 84.9 & 97.2 \\ 
&  F1 & 99.7 & 99.2 & 99.2 & 98.7 & 98.6 & 98.2 & 98.0 & 97.2 & 94.1 & 82.9 & 96.5 \\ 
 \hline

YOLOV8-X  & Precision & 99.0 & 99.0 & 98.8 & 98.8 & 98.8 & 98.7 & 97.7 & 96.8 & 94.0 & 89.5 & 97.1\\ 
 &  Recall & 99.8 & 99.8 & 99.6 & 99.3 & 99.3 & 99.1 & 98.4 & 98.0 & 95.3 & 92.0 & 98.1 \\ 
&  F1 & 99.4 & 99.4 & 99.2 & 99.0 & 99.0 & 98.9 & 98.0 & 97.4 & 94.6 & 90.7 & 97.6 \\ 
 \hline

FasterR-CNN & Precision & 97.9 & 97.8 & 96.8 & 96.8 & 95.6 & 94.5 & 94.5 & 93.5 & 89.7 & 76.0 & 93.3 \\ 
   &  Recall & 98.7 & 98.4 & 98.0 & 97.6 & 96.9 & 95.8 & 95.6 & 94.7 & 91.1 & 80.2 & 94.7 \\ 
 &  F1 & 98.3 & 98.1 & 97.4 & 97.2 & 96.2 & 95.1 & 95.0 & 94.1 & 90.4 & 78.0 & 94.0 \\ 
 \hline

MaskR-CNN & Precision & 98.9 & 97.8 & 97.7 & 96.5 & 96.4 & 95.4 & 95.4 & 94.5 & 91.2 & 71.2 & 93.5 \\ 
   &  Recall  & 99.3 & 98.9 & 98.7 & 98.0 & 97.6 & 96.9 & 96.2 & 95.8 & 92.7 & 76.6 & 95.1\\ 
 &  F1 & 99.1 & 98.3 & 98.2 & 97.2 & 97.0 & 96.1 & 95.8 & 95.1 & 91.9 & 73.8 & 94.3 \\ 
 \hline

TableDet & Precision & 98.5 & 97.5 & 97.5 & 97.4 & 96.3 & 95.3 & 94.4 & 93.5 & 90.7 & 77.2 & 93.8 \\ 
  &  Recall & 99.1 & 98.9 & 98.7 & 98.4 & 97.3 & 96.4 & 95.3 & 94.4 & 92.2 & 83.7 & 95.5 \\ 
 &  F1 & 98.8 & 98.2 & 98.1 & 97.9 & 96.8 & 95.8 & 94.8 & 93.9 & 91.4 & 80.3 & 94.6 \\ 
 \hline

DiffusionDet & Precision & 98.5 & 98.3 & 98.3 & 97.6 & 96.4 & 96.0 & 95.0 & 92.6 & 90.3 & 74.4 & 93.7\\ 
  &  Recall & 99.8 & 99.6 & 99.6 & 99.3 & 98.4 & 97.6 & 96.7 & 94.4 & 92.4 & 82.2 & 96.0 \\ 
 &  F1 & 99.1 & 98.9 & 98.9 & 98.4 & 97.4 & 96.8 & 95.8 & 93.5 & 91.3 & 78.1 & 94.8 \\ 
 \hline
 
Deformable- & Precision & 98.7 & 97.9 & 97.6 & 97.1 & 96.9 & 96.5 & 95.3 & 94.1 & 91.0 & 80.2 & 94.5\\ 
 DETR &  Recall & 99.6 & 99.3 & 99.1 & 98.9 & 98.9 & 98.7 & 97.8 & 96.7 & 94.4 & 86.6 & 97.0 \\ 
 &  F1 & 99.1 & 98.6 & 98.3 & 98.0 & 97.9 & 97.6 & 96.5 & 95.4 & 92.7 & 83.3 & 95.7 \\ 
 \hline
 
SparseR-CNN & Precision & 98.4 & 97.9 & 97.9 & 97.9 & 97.1 & 96.5 & 96.4 & 96.1 & 93.5 & 86.0 & 95.8\\ 
 &  Recall & 99.8 & 99.6 & 99.3 & 99.3 & 99.1 & 98.9 & 98.7 & 98.2 & 96.2 & 91.8 & 98.1 \\ 
 &  F1 & 99.1 & 98.7 & 98.6 & 98.6 & 98.1 & 97.7 & 97.5 & 97.1 & 94.8 & 88.8 & 96.9 \\ 
 \hline
SparseTableDet & Precision & 99.5 & 98.8 & 98.8 & 98.8 & 98.6 & 98.5 & 98.3 & 97.5 & 95.2 & 88.6 & 97.3 \\ 
(Proposed) & Recall & 100.0 & 99.8 & 99.6 & 99.6 & 99.6 & 99.6 & 99.6 & 98.9 & 97.3 & 92.0 & 98.6\\ 
 &  F1 & 99.7 & 99.3 & 99.2 & 99.2 & 99.1 & 99.0 & 98.9 & 98.2 & 96.3 & 90.3 & 97.9 \\ 
\end{longtable}

\begin{longtable}[c]{ c c c c c c c c c c c c c c}
\caption{Detailed Experimental results on the TNCR dataset.} \label{table:detailed_tncr_results} \\

\hline 
\multirow{2}{*}{Method} & \multirow{2}{*}{Metric} & \multicolumn{11}{{c}}{IoU}  \\ \cmidrule{3-13}
& &50\% &55\% &60\% &65\% &70\% &75\% &80\% &85\% &90\% &95\% & 50\%:95\% \\ \midrule
\endfirsthead

\multicolumn{13}{c}%
{{\tablename\ \thetable{} Detailed Experimental results on the TNCR dataset (continued from previous page).}} \\
\hline 
\multirow{2}{*}{Method} & \multirow{2}{*}{Metric} & \multicolumn{11}{{c}}{IoU}  \\ \cmidrule{3-13} 
& &50\% &55\% &60\% &65\% &70\% &75\% &80\% &85\% &90\% &95\% & 50\%:95\% \\ \midrule
\endhead

\hline \multicolumn{13}{r}{{Continued on next page}} \\ \hline
\endfoot

\hline 
\endlastfoot
 RetinaNet & Precision & 89.7 & 89.7 & 89.6 & 89.4 & 88.9 & 88.6 & 87.5 & 85.6 & 81.5 & 69.8 & 86.0\\ 
  &  Recall & 96.2 & 96.2 & 96.1 & 96.0 & 95.3 & 95.0 & 94.1 & 92.3 & 88.2 & 78.2 & 92.8 \\ 
 &  F1 & 92.8 & 92.8 & 92.7 & 92.6 & 92.0 &91.7 & 90.6 & 88.8 & 84.8 & 73.8 & 89.3\\ 
 \hline
  FCOS & Precision & 87.8 & 87.7 & 87.6 & 87.3 & 86.7 & 86.3 & 85.6 & 83.1 & 79.4 & 68.5 & 84.0\\ 
  &  Recall & 94.4 & 94.3 & 94.2 & 94.0 & 93.4 & 93.1 & 92.3 & 90.4 & 87.5 & 78.4 & 91.2 \\ 
 &  F1 & 91.0 & 90.9 & 90.8 & 90.5 & 89.9 & 89.6 & 88.8 & 86.6 & 83.3 & 73.1 & 87.5\\ 
 \hline
 YOLOX-X & Precision & 87.1 & 86.8 & 86.5 & 86.1 & 85.5 & 84.4 & 83.0 & 80.6 & 76.7 & 57.9 & 81.5\\ 
  &  Recall & 96.0 & 95.6 & 95.0 & 94.2 & 93.4 & 91.8 & 89.5 & 86.7 & 82.7 & 65.9 & 89.1 \\ 
 &  F1 & 91.3 & 91.0 & 90.6 & 90.0 & 89.3 & 87.9 & 86.1 & 83.5 & 79.6 & 61.6 & 85.1\\ 
 \hline

 YOLOR-X & Precision & 90.4 & 90.3 & 90.2 & 90.1 & 89.6 & 89.2 & 88.2 & 86.0 & 82.9 & 75.1 & 87.2\\ 
  &  Recall & 98.7 & 98.6 & 98.5 & 98.3 & 97.6 & 97.1 & 95.6 & 93.3 & 90.1 & 83.6 & 95.1 \\ 
 &  F1 & 94.4 & 94.3 & 94.2 & 94.0 & 93.4 & 93.0 & 91.8 & 89.5 & 86.4 & 79.1 & 91.0\\ 
 \hline

 YOLOV5-X & Precision & 93.0 & 92.8 & 92.7 & 92.4 & 92.2 & 91.8 & 91.1 & 88.9 & 86.0 & 79.3 & 90.0\\ 
  &  Recall & 99.3 & 99.2 & 99.1 & 99.0 & 98.9 & 98.5 & 97.9 & 96.1 & 93.4 & 88.2 & 97.0  \\ 
 &  F1 & 96.0 & 95.9 & 95.8 & 95.6 & 95.4 & 95.0 & 94.4 & 92.4 & 89.5 & 83.5 & 93.4\\ 
 \hline

YOLOV7-X & Precision & 92.0 & 91.9 & 91.8 & 91.7 & 91.6 & 91.3 & 90.3 & 88.6 & 85.8 & 77.3 & 89.2\\ 
  & Recall & 99.1 & 99.0 & 98.9 & 98.9 & 98.8 & 98.4 & 97.3 & 96.0 & 93.1 & 86.4 & 96.6 \\ 
 &  F1 & 95.4 & 95.3 & 95.2 & 95.2 & 95.1 & 94.7 & 93.7 & 92.2 & 89.3 & 81.6 & 92.8\\ 
 \hline

 YOLOV8-X & Precision & 93.1 & 93.1 & 93.0 & 92.8 & 92.4 & 92.2 & 91.4 & 89.5 & 86.6 & 79.2 & 90.3\\ 
  &  Recall & 99.3 & 99.3 & 99.3 & 99.2 & 98.9 & 98.6 & 98.0 & 96.3 & 93.9 & 87.6 & 97.0 \\ 
 &  F1 & 96.1 & 96.1 & 96.0 & 95.9 & 95.5 & 95.3 & 94.6 & 92.8 & 90.1 & 83.2 & 93.5\\ 
 \hline
 
FasterR-CNN & Precision & 89.1 & 89.1 & 88.9 & 88.7 & 88.5 & 88.2 & 87.8 & 86.5 & 81.5 & 66.6 & 85.5\\ 
  &  Recall & 94.3 & 94.3 & 94.2 & 94.1 & 93.7 & 93.4 & 93.0 & 91.7 & 87.5 & 76.1 & 91.2 \\ 
 &  F1 & 91.6 & 91.6 & 91.5 & 91.3 & 91.0 & 90.7 & 90.3 & 89.0 & 84.4 & 71.0 & 88.3\\ 
 \hline

MaskR-CNN & Precision & 90.2 & 90.0 & 90.0 & 89.8 & 89.7 & 89.1 & 88.2 & 86.3 & 81.7 & 64.4 & 85.9\\ 
  &  Recall & 95.3 & 95.2 & 95.1 & 95.0 & 94.8 & 94.3 & 93.7 & 92.0 & 88.0 & 75.5 & 91.9 \\ 
 &  F1 & 92.7 & 92.5 & 92.5 & 92.3 & 92.2 & 91.6 & 90.9 & 89.1 & 84.7 & 69.5 & 88.8\\ 
 \hline

TableDet & Precision & 91.7 & 91.7 & 91.7 & 91.5 & 91.3 & 90.7 & 90.2 & 88.4 & 84.0 & 72.9 & 88.4\\ 
  &  Recall & 98.1 & 98.1 & 98.0 & 97.9 & 97.7 & 97.0 & 96.6 & 95.2 & 91.7 & 83.6 & 95.4 \\ 
 &  F1 & 94.8 & 94.8 & 94.7 & 94.6 & 94.4 & 93.7 & 93.3 & 91.7 & 87.7 & 77.9 & 91.8\\ 
 \hline

Deformable- & Precision & 90.3 & 90.3 & 90.1 & 90.1 & 89.8 & 89.2 & 88.5 & 86.8 & 84.4 & 77.2 & 87.7\\ 
 DETR &  Recall & 99.2 & 99.1 & 99.0 & 98.9 & 98.8 & 98.6 & 97.9 & 97.0 & 94.7 & 89.4 & 97.3 \\ 
 &  F1 & 94.5 & 94.5 & 94.3 & 94.3 & 94.1 & 93.7 & 93.0 & 91.6 & 89.3 & 82.9 & 92.3\\ 
 \hline

 

DiffusionDet & Precision & 91.7 & 91.6 & 91.6 & 91.3 & 90.8 & 90.2 & 89.4 & 87.7 & 84.6 & 74.0 & 88.3 \\ 
  &  Recall & 99.6 & 99.6 & 99.6 & 99.4 & 98.7 & 97.9 & 97.2 & 95.5 & 92.9 & 85.2 & 96.6 \\ 
 &  F1 & 95.5 & 95.4 & 95.4 & 95.2 & 94.6 & 93.9 & 93.1 & 91.4 & 88.5 & 79.2 & 92.3\\ 
 \hline
 
 SparseR-CNN & Precision & 90.9 & 90.9 & 90.7 & 90.7 & 90.5 & 90.3 & 89.8 & 89.0 & 85.5 & 77.1 & 88.6 \\ 
  &  Recall & 99.9 & 99.8 & 99.8 & 99.7 & 99.7 & 99.7 & 99.5 & 98.7 & 97.1 & 89.9 & 98.4 \\ 
 &  F1 & 95.2 & 95.1 & 95.0 & 95.0 & 94.9 & 94.8 & 94.4 & 93.6 & 90.9 & 83.0 & 93.2\\ 
 \hline

SparseTableDet & Precision & 93.0 & 93.0 & 92.9 & 92.9 & 92.7 & 92.5 & 92.2 & 91.3 & 88.8 & 77.2 & 90.6 \\ 
(Proposed) &  Recall & 100.0 & 100.0 & 100.0 & 99.9 & 99.9 & 99.8 & 99.6 & 99.0 & 96.9 & 87.9 & 98.3\\ 
 &  F1 & 96.4 & 96.4 & 96.3 & 96.3 & 96.2 & 96.0 & 95.8 & 95.0 & 92.7 & 82.2 & 94.3 \\ 

\end{longtable}

\begin{longtable}[c]{ c c c c c c c c c c c c c c}
\caption{Detailed Experimental results on the ICT-TD dataset.} \label{table:icttd_detailed_results} \\

\hline 
\multirow{2}{*}{Method} & \multirow{2}{*}{Metric} & \multicolumn{11}{{c}}{IoU}  \\ \cmidrule{3-13}
& &50\% &55\% &60\% &65\% &70\% &75\% &80\% &85\% &90\% &95\% & 50\%:95\% \\ \midrule
\endfirsthead

\multicolumn{13}{c}%
{{\tablename\ \thetable{} Detailed Experimental results on the ICT-TD dataset (continued from previous page).}} \\
\hline 
\multirow{2}{*}{Method} & \multirow{2}{*}{Metric} & \multicolumn{11}{{c}}{IoU}  \\ \cmidrule{3-13} 
& &50\% &55\% &60\% &65\% &70\% &75\% &80\% &85\% &90\% &95\% & 50\%:95\% \\ \midrule
\endhead

\hline \multicolumn{13}{r}{{Continued on next page}} \\ \hline
\endfoot

\hline
\endlastfoot

RetinaNet & Precision & 95.8 & 95.7 & 94.7 & 94.4 & 92.4 & 91.1 & 90.0 & 87.9 & 82.1 & 68.0 & 89.2 \\ 
  &  Recall & 97.3 &   97.2 &   96.9 &   96.2 &   94.8 &   93.3 &   92.1 &   90.7 &   85.4 &   72.1 &   91.6 \\ 
 &    F1 &   96.5 &   96.4 &   95.8 &   95.3 &   93.6 &   92.2 &   91.0 &   89.3 &   83.7 &   70.0 &   90.4\\ 
 \hline
 
   FCOS &   Precision &   92.0 &   91.8 &   90.8 &   90.5 &   89.4 &   88.1 &   86.8 &   84.4 &   80.5 &   66.8 &   86.1 \\ 
 & Recall &   93.6 &   93.1 &   92.8 &   92.1 &   91.5 &   90.6 &   89.0 &   87.4 &   84.1 &   73.7 &   88.8 \\ 
 & F1 &   92.8 &   92.4 &   91.8 &   91.3 &   90.4 &   89.3 &   87.9 &   85.9 &   82.3 &   70.1 &   87.4\\ 
 \hline
 
  YOLOX-X &   Precision &   95.8 &   94.9 &   94.4 &   93.6 &   92.2 &   90.7 &   88.9 &   86.2 &   80.4 &   64.5 &   88.2 \\ 
  &    Recall &   98.7 &   97.8 &   97.3 &   96.7 &   95.1 &   93.5 &   91.4 &   88.3 &   82.9 &   67.9 &   91.0 \\ 
 &    F1 &   97.2 &   96.3 &   95.8 &   95.1 &   93.6 &   92.1 &   90.1 &   87.2 &   81.6 &   66.2 &   89.6\\ 
 \hline

   YOLOR-X &   Precision &   97.6 &   97.3 &   96.4 &   95.5 &   95.0 &   94.2 &   93.3 &   90.7 &   88.3 &   79.1 &   92.7 \\ 
  &    Recall &   99.4 &   99.1 &   98.6 &   97.9 &   97.1 &   96.5 &   95.3 &   92.1 &   89.8 &   81.1 &   94.7 \\ 
 &    F1 &   98.5 &   98.2 &   97.5 &   96.7 &   96.0 &   95.3 &   94.3 &   91.4 &   89.0 &   80.1 &   93.7\\ 
 \hline

   YOLOV5-X &   Precision &   97.4 &   97.2 &   97.2 &   97.0 &   96.4 &   95.6 &   94.8 &   93.7 &   90.7 &   81.2 &   94.1 \\ 
  &    Recall &   98.9 &   98.8 &   98.8 &   98.6 &   98.0 &   97.4 &   96.8 &   95.8 &   92.8 &   83.8 &   96.0 \\ 
 &    F1 &   98.1 &   98.0 &   98.0 &   97.8 &   97.2 &   96.5 &   95.8 &   94.7 &   91.7 &   82.5 &   95.0\\ 
 \hline

  YOLOV7-X &   Precision &   98.2 &   98.2 &   98.0 &   97.9 &   96.8 &   95.8 &   94.8 &   93.7 &   91.8 &   80.9 &   94.6 \\ 
  &    Recall &   99.5 &   99.4 &   99.3 &   99.1 &   98.5 &   97.8 &   96.7 &   95.6 &   93.4 &   83.5 &   96.3 \\ 
 &    F1 &   98.8 &   98.8 &   98.6 &   98.5 &   97.6 &   96.8 &   95.7 &   94.6 &   92.6 &   82.2 &   95.4\\ 
 \hline

  YOLOV8-X &   Precision &   97.9 &   97.1 &   97.0 &   96.9 &   96.4 &   95.6 &   94.7 &   93.6 &   91.4 &   82.4 &   94.3 \\ 
  &    Recall &   99.1 &   98.8 &   98.7 &   98.6 &   98.1 &   97.5 &   96.6 &   95.7 &   93.2 &   84.9 &   96.1 \\ 
 &    F1 &   98.5 &   97.9 &   97.8 &   97.7 &   97.2 &   96.5 &   95.6 &   94.6 &   92.3 &   83.6 &   95.2\\ 
 \hline
 
   FasterR-CNN &   Precision &   96.6 &   96.5 &   96.5 &   95.4 &   94.2 &   93.2 &   92.1 &   90.0 &   86.0 &   73.6 &   91.4\\ 
  &    Recall &   97.4 &   97.2 &   97.1 &   96.4 &   95.3 &   94.8 &   93.7 &   91.7 &   87.6 &   76.5 &   92.8 \\ 
 &    F1 &   97.0 &   96.8 &   96.8 &   95.9 &   94.7 &   94.0 &   92.9 &   90.8 &   86.8 &   75.0 &   92.1\\ 
 \hline

  MaskR-CNN &   Precision &   96.6 &   96.5 &   95.5 &   95.3 &   94.2 &   93.1 &   92.1 &   90.0 &   86.9 &   74.4 &   91.5 \\ 
  &    Recall &   97.4 &   97.1 &   96.9 &   96.3 &   95.5 &   94.4 &   93.4 &   91.7 &   88.9 &   77.8 &   92.9 \\ 
 &    F1 &   97.0 &   96.8 &   96.2 &   95.8 &   94.8 &   93.7 &   92.7 &   90.8 &   87.9 &   76.1 &   92.2\\ 
 \hline
 
 TableDet~\cite{xiao2023revisiting}  & Precision & 97.4 & 96.4 & 96.3 & 96.3 & 95.1 & 94.0 & 92.9 & 90.5 & 88.2 & 72.5 & 92.0 \\ 
 & Recall & 98.2 & 97.9 & 97.5 & 97.2 & 96.3 & 95.5 & 94.4 & 92.7 & 90.1 & 79.3 & 93.9\\ 
 &  F1 & 97.8 & 97.1 & 96.9 & 96.7 & 95.7 & 94.7 & 93.6 & 91.6 & 89.1 & 75.7 & 92.9 \\ 

\hline
DiffusionDet~\cite{xiao2023revisiting}  & Precision & 96.5 & 96.4 & 96.3 & 95.8 & 95.2 & 94.5 & 93.9 & 92.5 & 89.2  & 73.6 & 92.4 \\ 
 &   Recall & 99.3 & 99.2 & 99.0 & 98.8 & 98.4 & 97.8 & 97.2 & 96.0 & 93.1 & 79.5 & 95.8 \\ 
 &  F1 & 97.9 & 97.8 & 97.6 & 97.3 & 96.8 & 96.1 & 95.5 & 94.2 & 91.1 & 76.4 & 94.1 \\ 

\hline
Deformable-  & Precision & 97.0 & 96.6 & 96.3 & 96.0 & 95.2 & 94.4 & 93.7 & 92.6 & 89.7 & 80.3 & 93.2\\ 
DETR~\cite{xiao2023revisiting}& Recall & 98.9 & 98.6 & 98.5 & 98.3 & 97.8 & 96.9 & 96.3 & 95.2 & 92.8 & 85.8 & 95.9\\ 
 &  F1 & 97.9 & 97.6 & 97.4 & 97.1 & 96.5 & 95.6 & 95.0 & 93.9 & 91.2 & 83.0 & 94.5\\ 
 \hline
SparseR-CNN~\cite{xiao2023revisiting} & Precision & 96.2 & 96.0 & 95.5 & 95.3 & 94.2 & 93.3 & 92.6 & 91.1 & 88.3 & 75.6 & 91.8\\ 
 &  Recall & 99.0 & 98.9 & 98.7 & 98.4 & 97.7 & 97.0 & 96.2 & 94.9 & 92.5 & 82.4 & 95.6 \\ 
 &  F1 & 97.6 & 97.4 & 97.1 & 96.8 & 95.9 & 95.1 & 94.3 & 93.0 & 90.4 & 78.8 & 93.7 \\ 
 \hline
 SparseTableDet & Precision & 97.5 & 97.4 & 97.1 & 96.9 & 96.7 & 96.2 & 96.0 & 95.1 & 92.8  & 79.6 & 94.5\\ 
 (Proposed) &  Recall & 99.3 & 99.3 & 99.3 & 99.2 & 99.2 & 98.7 & 98.5 & 97.8 & 95.6 & 84.1 & 97.1 \\ 
 &  F1 & 98.4 & 98.3 & 98.2 & 98.0 & 97.9 & 97.4 & 97.2 & 96.4 & 94.2 & 81.8 & 95.8 \\ 
\end{longtable}





\end{document}